\documentclass{article} 
\usepackage{iclr2027_conference,times}

\usepackage{amsmath,amsfonts,bm}

\def\eqref#1{equation~\ref{#1}}
\def\1{\bm{1}}

\def\rd{{\textnormal{d}}}

\DeclareMathAlphabet{\mathsfit}{\encodingdefault}{\sfdefault}{m}{sl}
\SetMathAlphabet{\mathsfit}{bold}{\encodingdefault}{\sfdefault}{bx}{n}

\usepackage{algorithm}
\usepackage{algpseudocode}  
\usepackage{amsfonts}
\usepackage{amsmath}
\definecolor{mycite}{HTML}{0072BC}
\usepackage[colorlinks=true, citecolor=mycite, linkcolor=mycite, urlcolor=mycite]{hyperref}
\usepackage{url}
\usepackage{cleveref}
\usepackage{enumitem}
\usepackage{graphicx}
\usepackage{booktabs}
\usepackage{makecell}
\usepackage[table]{xcolor}
\usepackage{subcaption}
\usepackage{multirow}
\usepackage{wrapfig}
\usepackage[dvipsnames]{xcolor}

\usepackage{cleveref}
\crefname{appendix}{App.}{Apps.}
\Crefname{appendix}{App.}{Apps.}
\crefname{section}{Sec.}{Secs.}
\Crefname{section}{Sec.}{Secs.}
\crefname{equation}{Eq.}{Eqs.}
\Crefname{equation}{Eq.}{Eqs.}
\crefname{figure}{Fig.}{Figs.}
\Crefname{figure}{Fig.}{Figs.}
\crefname{table}{Tab.}{Tabs.}
\Crefname{table}{Tab.}{Tabs.}
\crefname{theorem}{Theorem}{Theorems}
\Crefname{theorem}{Theorem}{Theorems}
\crefname{proposition}{Proposition}{Propositions}
\Crefname{proposition}{Proposition}{Propositions}
\crefname{lemma}{Lemma}{Lemmas}
\Crefname{lemma}{Lemma}{Lemmas}
\crefname{remark}{Remark}{Remarks}
\Crefname{remark}{Remark}{Remarks}
\crefname{definition}{Definition}{Definitions}
\Crefname{definition}{Definition}{Definitions}
\crefname{observation}{Observation}{Observations}
\Crefname{observation}{Observation}{Observations}

\usepackage[most]{tcolorbox}
\usepackage{xcolor}
\definecolor{lavender}{RGB}{240, 240, 250}   % hex #F1F1FA
\definecolor{mint}{RGB}{229,242,229}  % hex #F1F8F7

\definecolor{gray}{RGB}{246,246,246}  % hex #F1F8F7
\definecolor{cfblue0}{HTML}{E8EFFF}  
\definecolor{cfblue1}{HTML}{e0eaff}  
\definecolor{cfblue2}{HTML}{c4d7ff}  
\definecolor{cfblue3}{HTML}{a4c1ff}  
\definecolor{cfblue4}{HTML}{7fa8ff}  
\definecolor{cfblue5}{HTML}{7499ff}  
\definecolor{cfblue6}{HTML}{1866B1}  
\definecolor{beige}{HTML}{FAF8F3} % beige
\definecolor{cfblue7}{HTML}{4652A8}  
\newcommand{\cfblue}[1]{{\color{cfblue7}#1}}

\newtcolorbox{plainbox_cfblue}{
    enhanced,
    breakable,
    colback=cfblue0,
    colframe=cfblue0,
    boxrule=0pt,
    left=3pt,
    right=3pt,
    top=5pt,
    bottom=5pt,
    before skip=5pt,
    after skip=5pt
}

\newtcolorbox{plainbox_mint}{
    enhanced,
    breakable,
    colback=mint,
    colframe=mint,
    boxrule=0pt,
    left=3pt,
    right=3pt,
    top=5pt,
    bottom=5pt,
    before skip=5pt,
    after skip=5pt
}

\newtcolorbox{plainbox_lavender}{
    enhanced,
    breakable,
    colback=lavender,
    colframe=lavender,
    boxrule=0pt,
    left=3pt,
    right=3pt,
    top=5pt,
    bottom=5pt,
    before skip=5pt,
    after skip=5pt
}

\newtcolorbox{plainbox_gray}{
    enhanced,
    breakable,
    colback=gray,
    colframe=gray,
    boxrule=0pt,
    left=10pt,
    right=10pt,
    top=2pt,
    bottom=2pt,
    before skip=3pt,
    after skip=10pt
}
\newtcolorbox{plainbox_beige}{
    enhanced,
    breakable,
    colback=beige,
    colframe=beige,
    boxrule=0pt,
    left=3pt,
    right=3pt,
    top=5pt,
    bottom=5pt,
    before skip=5pt,
    after skip=5pt
}
\definecolor{algbg}{HTML}{FFFCE0}
\definecolor{algcomment}{HTML}{005E0B} % green comment
\definecolor{algkw}{HTML}{4B3F8C}
\definecolor{algfunc}{HTML}{2E75B6}
\definecolor{algemph}{HTML}{9B2C2C}

\lstdefinestyle{idf}{
  language=Python,
  backgroundcolor=\color{algbg},
  basicstyle=\ttfamily\scriptsize,
  commentstyle=\color{algcomment},
  keywordstyle=\color{algkw}\bfseries,
  emph={randn,rerope,range,len,append},
  emphstyle=\color{algfunc},
  emph={[2]G,E},
  emphstyle={[2]\color{algemph}\bfseries},
  showstringspaces=false,
  columns=fullflexible,
  xleftmargin=8pt,
  framexleftmargin=8pt,
  aboveskip=4pt, belowskip=4pt,
}
\usepackage{titletoc}
\usepackage[most]{tcolorbox}
\usepackage{kotex}
\usepackage{relsize}
\newcommand{\emailfont}[1]{{\mdseries\relscale{0.95}\texttt{#1}}}
\usepackage{listings}

\title{Diverse Motion Customization via \\ Control-based Dynamic Optimization}
\author{
\textbf{Youngyoon Choi}$^{1*}$, \quad
\textbf{Kihyun Kim}$^{1,2*}$, \quad
\textbf{Jeongwoo Shin}$^{1\dagger}$, \quad
\textbf{Joonseok Lee}$^{1\dagger}$ \\
{\mdseries $^{1}$Seoul National University, \ $^{2}$AIM Intelligence} \\
\emailfont{\{youngyoon911, ki5477, swswss, joonseok\}@snu.ac.kr}
}

\iclrfinalcopy % Uncomment for camera-ready version, but NOT for submission.
\begin{document}

\maketitle
{\renewcommand{\thefootnote}{}%
 \footnotetext{$^{*}$Equal contribution. \quad $^{\dagger}$Corresponding author.}}
\vspace{-.8cm}
\begin{figure}[H]
    \centering
    \includegraphics[width=.9\linewidth]{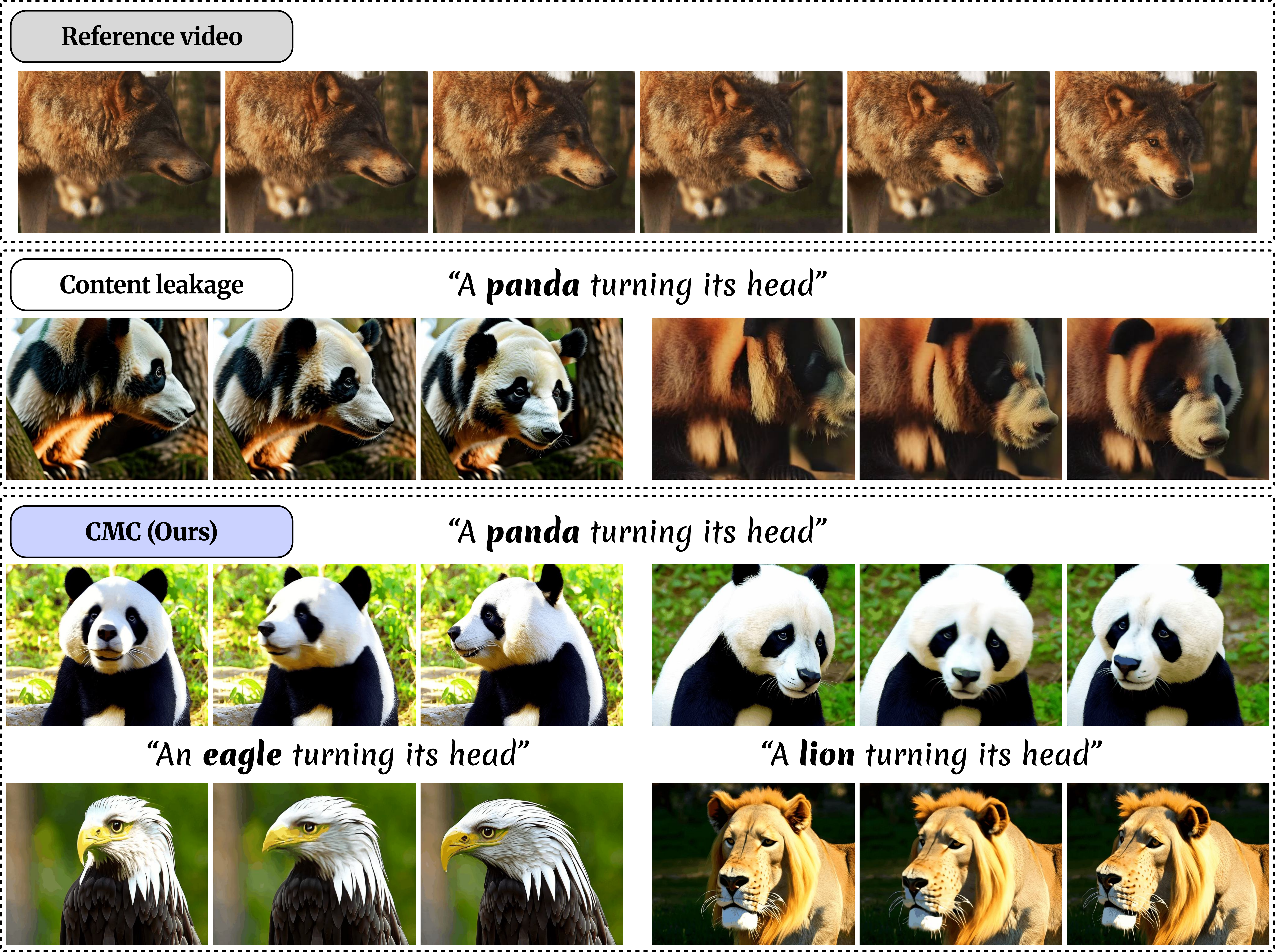}
    % \vspace{-.4cm}
    \caption{\textbf{Overview of the proposed method.} Our CMC mitigates content leakage, enabling diverse generation across varying prompts and samples.}
    \label{fig:overview}
    \vspace{-.4cm}
\end{figure}

\begin{abstract}
Despite recent advances in video generation, motion customization remains challenging due to \emph{content leakage}, where appearance attributes from the reference video unintentionally propagate into the generated output. We identify this issue as a consequence of the generative process collapsing toward the reference video, which arises from formulating the learning objective as a direct regression on the reference. To address this, we propose \textbf{Control-based Motion Customization (CMC)}, a principled training framework that is structurally robust to content leakage. Our key idea is to steer generative dynamics toward desired motion while avoiding collapse toward the reference video, which we formalize using Stochastic Optimal Control (SOC). Under this formulation, customized videos acquire the target motion yet remain within the pre-trained model's prompt-conditional distribution, where appearance is determined by the text prompt rather than the reference video.
Furthermore, to improve efficiency, we tailor the SOC formulation to motion customization by eliminating the need for an explicit reward and introducing a timestep-adaptive motion cost that focuses only on early generative stages, accelerating training by $2.5\times$. Extensive experiments demonstrate that CMC effectively mitigates content leakage and achieves competitive motion fidelity while preserving the diversity of the base model across diverse scenarios.
\end{abstract}

% --- Main ---
\section{Introduction}
\label{sec:intro}

Recent video generation models~\citep{blattmann2023align, chen2024videocrafter2, guo2024animatediff, gupta2023photorealistic, wang2024lavie} have demonstrated remarkable progress in visual quality and temporal coherence, bringing video synthesis closer to practical real-world applications.
Among these, motion customization has emerged as an important task to provide users with finer control over the generation process.
In particular, motion customization aims to generate a video conditioned on (i) a text prompt specifying the overall scenario, \emph{e.g.}, subject appearance or background details, and (ii) a reference video specifying the target motion.
Thus, the output should follow the motion of the reference video, while matching all other attributes specified in the text prompt.

However, motion customization still remains challenging due to its trade-off between motion fidelity and \emph{content leakage}, where appearance attributes from the reference video 
unintentionally propagate into the generated result~\citep{maeffivmt, kansy2025reenact, liu2025motionshot}.
Existing methods commonly attempt to mitigate this by debiasing appearance from motion representations extracted from the reference video, yet this is inherently difficult.
We identify that content leakage does not stem from imperfect disentanglement itself, but from formulating the learning objective as a \emph{direct regression} on the reference video. 
This causes the generation to drift away from the text prompt toward the reference video, making the output resemble the reference beyond its motion. Based on this observation, we argue that content leakage can be better mitigated simply by designing a proper learning objective, even when motion and appearance remain entangled.

To this end, we introduce \textbf{Control-based Motion Customization (CMC)}, a principled training framework for motion customization. Our goal is to adopt only the target motion while faithfully reflecting the text prompt, which requires preventing collapse toward the reference. To achieve this, we first highlight the importance of rigorously accounting for the generative dynamics and their induced output distribution (\S\ref{sec:method_4.1}). We then adopt Stochastic Optimal Control (SOC) as a natural tool for this purpose, casting motion customization as an SOC problem (\S\ref{sec:method_4.2}).
Specifically, instead of forcing the model to match the reference directly, CMC learns a control that steers the pre-trained generative process toward the target motion, while being penalized for deviating from the original process. The customized samples thus acquire the desired motion yet remain within the pre-trained prompt-conditional distribution: since appearance is determined by the prompt, samples from this distribution are unlikely to exhibit appearance attributes inconsistent with the prompt, including those of the reference video. Consequently, CMC is structurally robust to content leakage.

While the SOC framework offers a principled approach, its standard application remains challenging for two practical reasons.
First, it requires carefully handcrafted reward functions that are difficult to define for complex motion customization task.
Second, it is computationally expensive due to its full-horizon trajectory simulation.
To address these challenges, we redesign the formulation through two task-specific modifications.
First, we completely eliminate the explicit reward.
Instead, we modify the generative dynamics using a proxy objective computed along the trajectory, showing that the resulting target distribution still remains as a reweighting of the pre-trained conditional distribution (\S\ref{sec:method_4.3}).
Second, based on our observation that motion structure emerges early in generative stages, we restrict this dynamics modification to these initial timesteps (\S\ref{sec:method_4.4}).
These two task-oriented modifications simplify the objective, leading to a more efficient control-learning process (\S\ref{sec:method_4.5}). 
Extensive experiments demonstrate that our framework enables precise motion control while remaining robust to content leakage, thereby facilitating diverse generation across a wide range of scenarios. 

Our contributions are summarized as follows:
\begin{itemize}[leftmargin=2em,topsep=0.3em,parsep=0pt,partopsep=0pt]
    \item We identify the inherent susceptibility of prevalent motion customization methods to content leakage from the perspective of generative dynamic and induced target distribution.
    \item We propose \textbf{Control-based Motion Customization (CMC)}, an SOC-based framework that transfers the target motion while remaining faithful to the text prompt, making it structurally robust to content leakage.
    \item We tailor the formulation to motion customization by (i) removing the reward and directly guiding the generation toward the target motion, and (ii) restricting this guidance to early generative stages, accelerating training by $2.5\times$.
    \item Through comprehensive experiments, we verify that our method yields \textit{diverse} generation results with \textit{high motion fidelity}, effectively \textit{mitigating content leakage}.
\end{itemize}
\section{Related Work}
\label{sec:rel_work}

\textbf{Training-Free Motion Customization.}
Training-free approaches~\citep{hu2024motionmaster,meral2024motionflow, liu2025motionshot, zhang2025motionconsistency, yatim2024, ling2024motionclone, gao2025conmo, pondaven2025video, yesiltepe2024motionshop, gokmen2025ropecraft} customize motions by manipulating sampling dynamics at inference time, without updating model parameters.
Early approaches \citep{hu2024motionmaster, meral2024motionflow, liu2025motionshot, zhang2025motionconsistency, yatim2024, ling2024motionclone, gao2025conmo} are based on
U-Net \citep{ronneberger2015unet}, leveraging its intermediate representations to align generation with the referred motion,
while recent approaches \citep{pondaven2025video, yesiltepe2024motionshop, gokmen2025ropecraft} adopt Diffusion Transformers (DiT)~\citep{peebles2023scalable}.
Despite architectural differences, these methods commonly enforce motion alignment by directly modifying latent representations or internal features.
In contrast, our method operates at the level of generative dynamics without explicitly altering intermediate representations.

\textbf{Fine-tuning-based Motion Customization.}
Another line of work transfers motion by fine-tuning model parameters, often from a single reference~\citep{ kansy2025reenact,park2025spectral,wang2025motion,shi2025decouple}. Many of these approaches introduce separate appearance and motion branches to encourage disentangled representations~\citep{zhao2024motiondirector, zhang2023motioncrafter, ren2024customize, maeffivmt, liu2025separate}. Others adapt text embeddings and attention layers to better inject motion cues~\citep{li2024motrans, materzynska2024newmove, wu2025motionmatcher,jeong2024vmc}. While effective, these methods often suffer from output distribution collapse toward the reference video. 
Beyond single-exemplar adaptation, large-scale fine-tuning approaches~\citep{ressler2025dismo, cai2025efficientmt, zhang2025flexiact} learn motion control from curated datasets. Such approaches rely heavily on large annotated data, whereas we focus on the single-reference setting without large-scale training.
\section{Preliminary}
\label{sec:preliminary}

\textbf{Text-to-Video Diffusion.}
Let us consider a continuous-time stochastic differential equation (SDE) formulation of text-to-video diffusion and flow models~\citep{blattmann2023align, chen2024videocrafter2, guo2024animatediff, gupta2023photorealistic, wang2024lavie, wan2025wan}.
Given a data point $X_1 \in \mathbb{R}^n$ sampled from data distribution $q_{\text{data}}$, 
a forward stochastic process gradually transforms the data into Gaussian distribution {$q_{\text{prior}}$}.
Using the reversed time convention, the forward SDE is given by
\begin{equation}
\rd X_t = f(X_t, t)\rd t + \sigma(t)\rd W_t,
\qquad X_1 \sim q_\text{data},
\label{eq:preliminary:forward_sde}
\end{equation}
\noindent
where $f: \mathbb{R}^n \times [0,1] \rightarrow \mathbb{R}^n$ is a forward drift, $\sigma:[0,1] \rightarrow \mathbb{R}_{>0}$ is a diffusion coefficient, and $W_t$ is an $n$-dimensional standard Wiener process.
Both $f$ and $\sigma$ are determined by a predefined noise schedule.

Reversing \cref{eq:preliminary:forward_sde} yields a reverse-time stochastic process that transports 
$q_{\text{prior}}$
back to the 
$q_{\text{data}}$:
\begin{equation}
\rd X_t = \big[f(X_t, t) + {\sigma(t)}^2 s^{\text{base}}(X_t,t)\big]\rd t + \sigma(t)\rd W_t, \quad X_0 \sim q_{\text{prior}},
\label{eq:preliminary:backward_sde}
\end{equation}
where score function $s^{\text{base}}:\mathbb{R}^n \times [0,1] \rightarrow \mathbb{R}^n$ is typically parameterized and optimized by conditional score matching~\citep{song2021scorebased} to estimate the marginal score $\nabla_{x} \log p(x,t)$;
for flow models, it is obtained in closed-form from the learned velocity field~\citep{sit}.
Once trained, 
% the learned score function 
it specifies the reverse-time dynamics, enabling sampling by simulating
the reverse SDE~\cref{eq:preliminary:backward_sde} from 
$q_{\text{prior}}$ at $t=0$
to obtain samples $X_1$ that approximately follow the 
$q_{\text{data}}$ at $t=1$. Also, it induces a path measure $p^\text{base}$ with endpoint marginals $p^\text{base}_0=q_{\text{prior}}$ and $p^\text{base}_1\approx q_{\text{data}}$.

\textbf{Stochastic Optimal Control (SOC).}
SOC provides a framework for modifying the dynamics of a given stochastic process to arrive at a new target distribution $\mu$ at $t=1$ by introducing a control signal that manipulates trajectories,  
while remaining close to a base path measure~\citep{zhang2021pis,vargas2023dds,domingo2024am,havens2025as,liu2025asbs,shin2026asbm}. 
Among the various SOC problem formulations, we consider the quadratic cost control-affine one:
% \newpage
\begin{equation}
\label{eq:preliminary:soc_obj}
\min_{u \in \mathcal{U}}\ \mathbb{E}_{X\sim p^{u}}\!\left[
\int_{0}^{1} \left(\frac{1}{2}\,\|{u}(X_t,t)\|^{2}\, + \ell(X_t, t) \right)\rd t+ g(X_1)
\right]
\end{equation}
\begin{equation}
\text{s.t.} \,\,\, \rd X_t = \big[b(X_t,t) + \sigma(t) {u}(X_t, t)\big]\,\rd t + \sigma(t)\,\rd W_t, \,\,\
X_0 \sim q_{\text{prior}}, 
\label{eq:preliminary:soc_sde}
\end{equation}

\noindent
where $b: \mathbb{R}^n \times [0,1] \rightarrow \mathbb{R}^n$ is a base backward drift, $u: \mathbb{R}^n \times [0,1] \rightarrow \mathbb{R}^n$ is a control vector field ranging over the set of admissible controls $\mathcal{U}$, $\ell: \mathbb{R}^n \times [0,1] \rightarrow \mathbb{R}$ is a running cost and $g: \mathbb{R}^n \rightarrow \mathbb{R}$ is a terminal cost. 
By optimizing \cref{eq:preliminary:soc_obj}, we can obtain minimally changed dynamic which leads to our new target distribution $\mu$ at $t=1$ (via terminal cost $g(x)$), under additional constraints on trajectory shaping (via running cost $\ell(x,t)$).

\begin{figure}[t]
    \centering
    \includegraphics[width=1\linewidth]{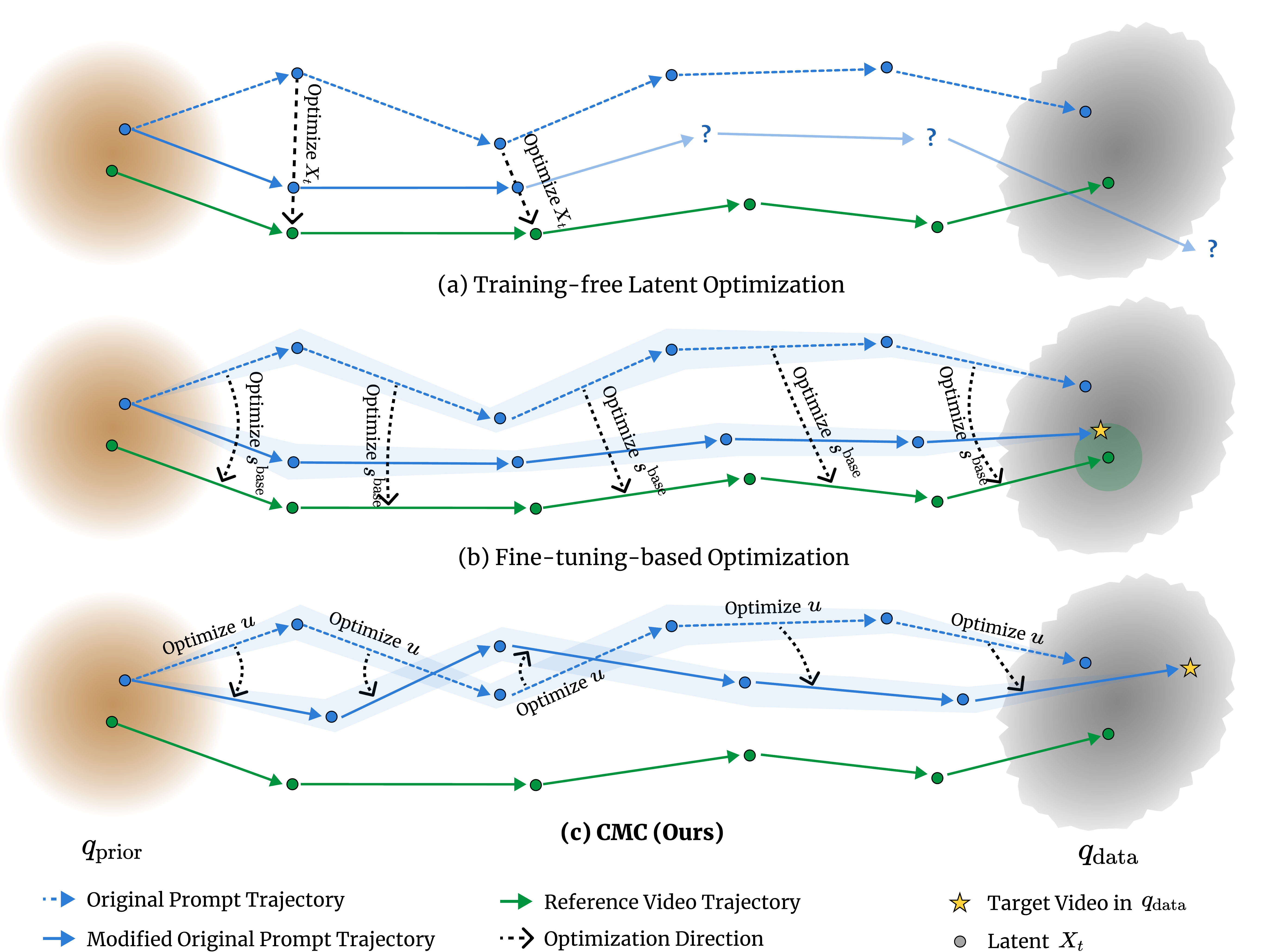}
    \vspace{-.4cm}
    \caption{\textbf{Method Comparison.} Existing methods \textit{vs}. our control-based approach.}
    \label{fig2:method}
    \vspace{-.4cm}
\end{figure}

\section{Control-based Motion Customization (CMC)}
\subsection{Motivation}
\label{sec:method_4.1}
We begin by revisiting existing motion customization methods through the lens of generative dynamics and formulate our problem based on this analysis.

\textbf{Reinterpreting Existing Methods.} Training-free approaches~\citep{yatim2024, gao2025conmo, ling2024motionclone, pondaven2025video} encode motion by adjusting the intermediate noisy state $X_t$ to be closer to the generative trajectory of the reference video.
However, such direct alignment is inherently susceptible to content leakage, as its minimizer is precisely the generative trajectory of the reference video. Since no constraint prevents the model from converging to this trajectory, and the induced output distribution is left uncharacterized, the output can easily collapse toward the reference video.
\Cref{fig2:method}(a) visualizes how latent-level alignment progressively bends the original trajectory toward the reference, disrupting the intrinsic generative dynamics.

Fine-tuning approaches~\citep{zhao2024motiondirector, maeffivmt, jeong2024vmc, park2025spectral, wang2025motion, wu2025motionmatcher}, on the other hand, directly optimize the parameters of the pre-trained model, effectively updating the score function $s^{\text{base}}(x,t)$ in the backward dynamics~\cref{eq:preliminary:backward_sde}.
During training, the model reconstructs the reference video via diffusion loss, which leads original trajectories to converge toward the reference trajectory as in \cref{fig2:method}(b).
While this ensures that the generated samples lie in $q_{\text{data}}$, it causes the output distribution to collapse into a narrow point-mass-like distribution centered at the reference, resulting in content leakage and limited flexibility.
\subsection{Motion Customization as SOC}
\label{sec:method_4.2}
Aforementioned observations motivate a training framework that rigorously accounts for both the \cfblue{\emph{generative dynamics}} and the induced \cfblue{\emph{output distribution}}.
To this end, we formulate motion customization as a SOC problem.
As shown in the SOC objective in \cref{eq:preliminary:soc_obj}, this formulation enables us to incorporate the desired motion information by learning control $u$ while (i) preventing excessive deviation from the base dynamics through the quadratic cost $\|u\|^2$, and (ii) explicitly controlling the induced output distribution $\mu$.

Adopting this SOC framework leads to the following optimization problem for obtaining the optimal control $u^\star$, subject to our controlled generative dynamics:
\begin{equation}
\rd X_t = \big[f(X_t, t)+ {\sigma(t)}^2s^{\text{base}}(X_t, t) + \sigma(t) {u}(X_t, t)\big]\rd t + \sigma(t)\rd W_t, \,\, X_0 \sim q_{\text{prior}},
\label{eq:method:our_sde}
\end{equation}
\noindent
which induces the controlled path measure $p^{u}$ that coincides with the base path measure $p^{\text{base}}$ when $u \equiv 0$. Within this objective, the target motion information can be flexibly injected either through the target distribution $\mu$ or via a specifically designed running cost $\ell(x,t)$.

\textbf{Challenges of Standard SOC Application.}
A natural way to learn the desired motion within this framework
is to specify the target distribution $\mu$ as the distribution of videos containing the target motion. In practice, this distribution is commonly formulated using a reward function
$r: \mathbb{R}^n \rightarrow \mathbb{R}$, which defines $\mu$ as a reward-tilted distribution:
\begin{equation}
    \mu(x) \propto p_1^{\text{base}}(x)\exp(r(x)).
\label{eq:method:tilted_dist}
\end{equation}
Under this setting, the terminal cost in \cref{eq:preliminary:soc_obj} corresponds to a terminal reward, \emph{i.e.}, $g(X_1) = -r(X_1)$.
However, directly optimizing this objective with terminal reward $r(X_1)$ is practically challenging and computationally expensive. Since $r$ explicitly defines the target distribution, it must capture multiple aspects of the generated video, including motion alignment, visual quality, and text consistency. Yet, designing such a comprehensive reward is nontrivial~\citep{xue2025dancegrpo, liu2026beyond, liuflow}. 
Furthermore, evaluating $r(X_1)$ requires simulating the entire trajectory up to $t=1$, which substantially increases optimization costs. To overcome these inefficiencies, we reformulate the SOC objective specifically for motion customization, yielding our framework, \textbf{Control-based Motion Customization (CMC)}. Specifically, CMC is designed to achieve the following two goals:

\begin{plainbox_beige}
\begin{itemize}[label={}, itemsep=3pt, topsep=0pt, parsep=0pt, partopsep=0pt, leftmargin=*]
    \item \textbf{Goal 1:} Inject the target motion \emph{without} explicit reward design (\S\ref{sec:method_4.3}).
    \item \textbf{Goal 2:} Inject the target motion \emph{without} entire trajectory simulation (\S\ref{sec:method_4.4}).
\end{itemize}
\end{plainbox_beige} 
\begin{figure}[t]
    \centering
    \includegraphics[width=1\linewidth]{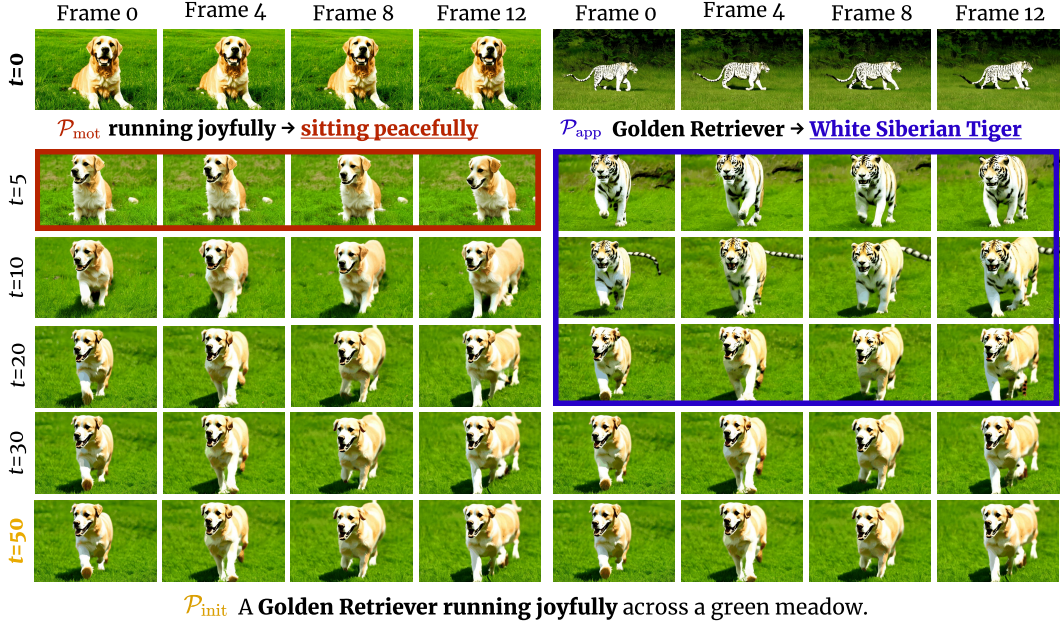}
    \caption{
    \textbf{Qualitative analysis on motion formation timestep.}
    The $y$-axis indicates the \textbf{switching timestep} ($t\in[0,50]$) at which the initial prompt $\mathcal{P}_{\text{init}}$ is switched to a new prompt $\mathcal{P}_{\text{mot}}$ or $\mathcal{P}_{\text{app}}$. 
    \emph{Left}: Switching from $\mathcal{P}_{\text{init}}$ to $\mathcal{P}_{\text{mot}}$ shows that motion becomes largely established after early timesteps ($t=10$), with limited influence from subsequent prompts. \emph{Right}: Switching to $\mathcal{P}_{\text{app}}$ shows that non-motion  
    can still change at later timestep ($t=20$).}
    \label{fig3:early_qual}
    \vspace{-8px}
\end{figure}

\subsection{Bypassing Reward Design via Zero Terminal Cost}
\label{sec:method_4.3}
Our first main insight is that we can entirely bypass the complexity of reward design by avoiding the need to construct a new, custom target distribution $\mu$. As modern text-to-video diffusion models \citep{yangcogvideox, wan2025wan, kong2024hunyuanvideo, zheng2024open} are trained on vast and diverse datasets, their pre-trained output distribution already captures rich generative knowledge. 
Under the natural assumption that the desired motion lies within this underlying distribution, 
we set the terminal cost $g(x)=0$ to avoid explicitly designing the target distribution as a reward-tilted distribution (\cref{eq:method:tilted_dist}), and instead utilize a running cost $\ell(x,t)$ to \emph{implicitly} modify the dynamics.

The target distribution induced by $\ell$ can be explicitly characterized. Since motion customization is fundamentally a form of text-to-video generation, $p_1^{\mathrm{base}}$ denotes the \emph{prompt-conditional} distribution $p_1^{\mathrm{base}}(\cdot \mid c)$ given a text prompt $c$, and we make this conditioning explicit hereafter.

% \newpage
\vspace{-.2cm}
Under this formulation, our optimal target distribution is derived as:
\begin{equation}
p_1^\star(x \mid c) = p_1^{\mathrm{base}}(x \mid c)\, \phi(x),
\quad \phi(x) = \mathbb{E}_{p^{\mathrm{base}}(\cdot \mid c)}\Big[e^{-\int_0^1 \ell(X_t,t)\,\mathrm{d}t + V(X_0,0)} \,\Big|\, X_1 = x\Big],
\label{eq:method:reweighted_dist}
\end{equation}
where $V(x,0) = -\log \mathbb{E}_{p^{\mathrm{base}}}[e^{-\int_0^1 \ell(X_t,t)\,\mathrm{d}t} \mid X_0 = x]$. Derivation is in Appendix \S\ref{sec:appendix:soc}.

Importantly, since the optimal target distribution is a \emph{reweighting} of the base terminal distribution, our formulation is inherently robust to content leakage.

\vspace{.2cm}
\begin{plainbox_gray}
\textbf{Property 1: Support Preservation.}
\textit{$p_1^\star(\cdot \mid c)$ cannot assign probability to samples that do not exist under $p_1^{\mathrm{base}}(\cdot \mid c)$.}
\end{plainbox_gray}
Conditioning on $c$ implies that the appearance of the generated video is determined by the prompt.
Accordingly, samples exhibiting content leakage, \emph{i.e.}, appearance attributes from the reference video that contradict the prompt $c$, are already assigned negligible probability under $p_1^{\mathrm{base}}$. Since reweighting \cref{eq:method:reweighted_dist} is pointwise on $p_1^{\mathrm{base}}$, content-leaked samples with $p_1^{\mathrm{base}}(x) \approx 0$ cannot be newly introduced through this process. This property is further supported by the quadratic control regularization, which prevents the point-wise scaling factor $\phi(x)$ from exploding.

\vspace{.2cm}
\begin{plainbox_gray}
\textbf{Property 2: Trajectory Regularization}.
\textit{The induced path measure $p^u$ cannot deviate arbitrarily from the base path measure $p^{\mathrm{base}}$, due to the regularization imposed by the control cost.}
\end{plainbox_gray}
Complementing the sample-level bound in Property 1, the quadratic control cost enforces trajectory-level bound to the base model. Because the control cost directly equals the KL divergence between path measures~\citep{sarkka2019appliedSDE}, $\mathbb{E}_{p^u}[\frac{1}{2}\int_0^1 \Vert{}u\Vert{}^2\,\mathrm{d}t] = \mathrm{KL}(p^u \,\Vert{}\, p^{\mathrm{base}})$, minimizing $\Vert{}u\Vert{}^2$ keeps the 
modified dynamics strictly anchored to $p^{\mathrm{base}}$ throughout the entire generation process, preventing intermediate states from drifting toward unverified reference features.

\begin{plainbox_beige}
\begin{itemize}[label={}, itemsep=3pt, topsep=0pt, parsep=0pt, partopsep=0pt, leftmargin=*]
    \item \textbf{Solution 1: Vanishing Terminal Cost.} Inject the target motion \emph{via} the running cost $\ell(x,t)$, with $g(x) = 0$.
\end{itemize}
\end{plainbox_beige} 
\subsection{Trajectory Truncation via Early-Stage Control}
\label{sec:method_4.4}
While our formulation operates via a running cost $\ell(x, t)$, evaluating this cost across all denoising steps induces unnecessary optimization overhead. Therefore, we leverage the key property that motion structure primarily emerges during the early denoising stage, and restrict control to an early time window $t \in [0, \tau]$, achieving both computational efficiency and high generative flexibility.

\textbf{Motion Formation in Early Diffusion Steps.} 
Inspired by prior observations~\citep{ling2024motionclone,hertz2022prompt} and the coarse-to-fine nature of diffusion models~\citep{yi2024towards,qian2024boosting}, we hypothesize that motion structure is largely established in \emph{early} denoising steps. To validate this hypothesis, we conduct both qualitative and quantitative analyses.

For qualitative analysis, we conduct a prompt-switching experiment. We initiate generation with $\mathcal{P}_{\text{init}}$ and switch to either a motion-changed prompt $\mathcal{P}_{\text{mot}}$ (same subject, different motion) or an appearance-changed prompt $\mathcal{P}_{\text{app}}$ (different subject, same motion).
As shown in \cref{fig3:early_qual}(\emph{left}), motion is largely shaped within approximately the first 20\% of the trajectory; switching the prompt afterward leaves the motion nearly unchanged.
In contrast, in \cref{fig3:early_qual}(\emph{right}), appearance remains modifiable at substantially later timesteps (first $40\%$), allowing the subject to adapt to the new description while motion persists.

\begin{wrapfigure}{r}{0.4\textwidth}
    \centering
    \vspace{-.4cm}
    \includegraphics[width=\linewidth]{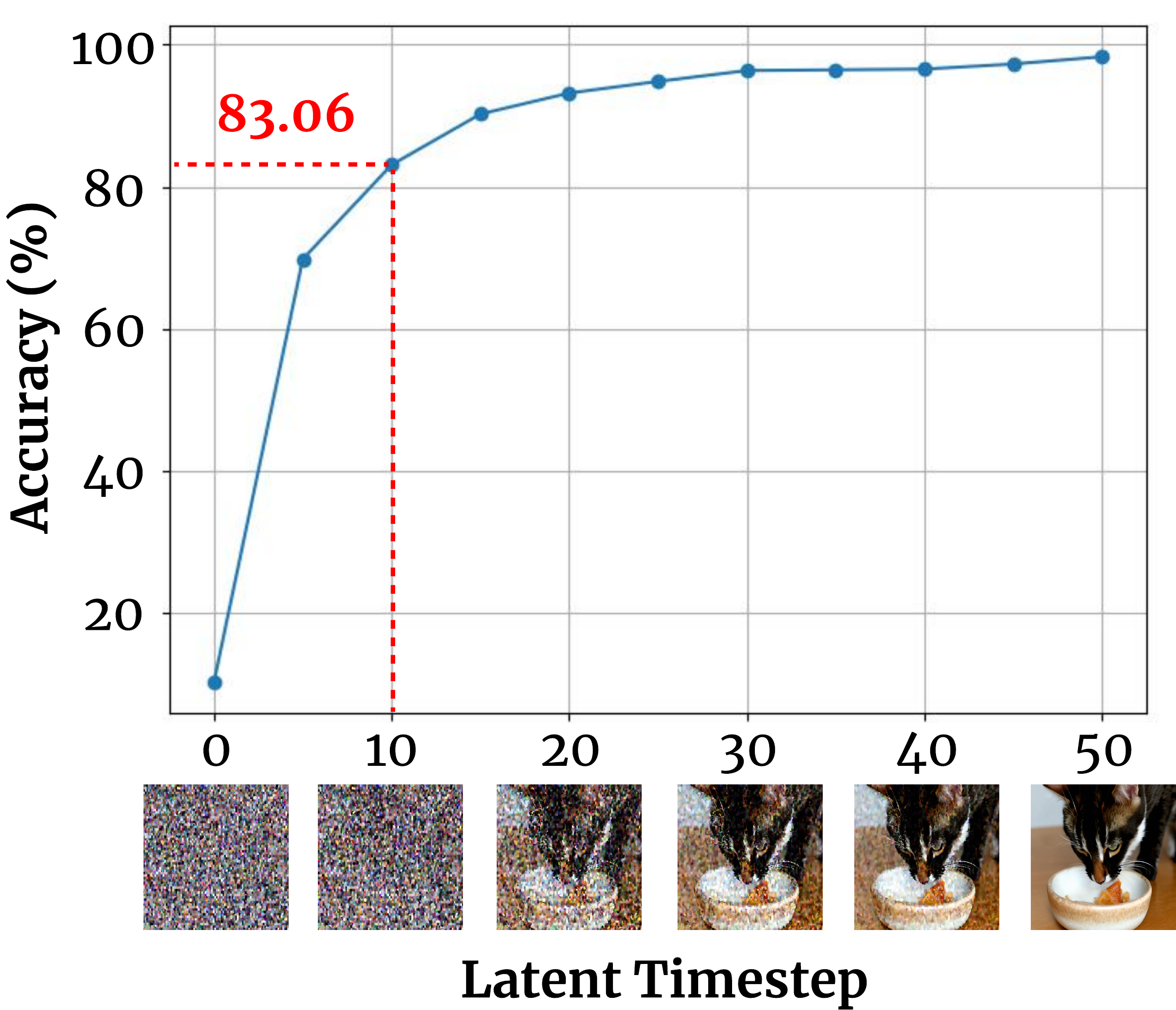}
    \vspace{-.4cm}
    \caption{Quantitative analysis on motion formation timestep.}
    \vspace{-.4cm}
    \label{fig4:early_quant}
\end{wrapfigure}

For quantitative analysis, we evaluate a motion classifier trained on intermediate noisy latents $X_t$ across a benchmark of 10 motion classes and 10 subject categories (details in Appendix \S\ref{sec:appendix:imple_details}). 
As shown in \cref{fig4:early_quant}, motion is highly identifiable from a very early stage (step 5 with 69.8\%) and accuracy saturates around step 15 (90.2\%), confirming that the primary motion structure is formed early in the denoising process.
Based on these findings, we conclude that imposing motion information only in the early generative phase is sufficient for motion control, resulting in a significantly more efficient control learning procedure. We therefore restrict the running cost $\ell(x,t)$ to $t \in [0,\tau]$, where $\tau < 1$ denotes the cutoff timestep for motion control. This restriction also preserves flexibility for subsequent semantic refinement, allowing the generation to adapt to the text prompt.

\textbf{Timestep-adaptive Motion Cost.} While the running cost $\ell(x,t)$ can be flexibly designed to encourage desired motions, we adopt a simple and effective instantiation computing optical-flow-like motion vectors directly in the feature space \citep{geyer2023tokenflow, wang2024cove}. Specifically, we formulate the cost as:
\begin{equation}
\ell(X_t, t) = w(t) \cdot \ell_{\mathrm{motion}}(X_t; X_{\mathrm{ref}}),
\label{eq:method:running_cost_inst}
\end{equation}
where $\ell_{\mathrm{motion}}$ quantifies the feature-space motion vector discrepancy relative to the clean reference video $X_{\mathrm{ref}}$, and $w(t)$ is a timestep-dependent scaling factor. All formal definitions are in Appendix \S\ref{sec:appendix:imple_details}. To further demonstrate that leakage prevention is structural to our framework rather than tied to a specific motion representation, we also instantiate $\ell(x,t)$ with motion costs adopted from existing methods (Appendix \S\ref{sec:appendix:more_results}).

\begin{plainbox_beige}
\begin{itemize}[label={}, itemsep=3pt, topsep=0pt, parsep=0pt, partopsep=0pt, leftmargin=*]
    \item \textbf{Solution 2: Running Cost Truncation.} Inject the target motion \emph{via} the early generative stage only, with $\ell(x,t) = 0$ for all $t \in (\tau, 1]$.
\end{itemize}
\end{plainbox_beige} 
\subsection{Training Strategy}
\label{sec:method_4.5}

By applying \textbf{Solution 1} ($g(x) = 0$) and \textbf{Solution 2} ($\ell(X_t,t) = 0 \quad \forall t \in (\tau, 1]$) to our general SOC objective in \cref{eq:preliminary:soc_obj}, the control problem simplifies to optimizing the truncated running cost over $t \in [0, \tau]$:
\vspace{-.2cm}
\begin{equation}
\label{eq:method:final_obj}
\min_{u \in \mathcal{U}}\ \mathbb{E}_{X\sim p^{u}}\!\left[
\int_{0}^{\tau} \left(\frac{1}{2}\,\|{u}(X_t,t)\|^{2}\, +  w(t) \cdot \ell_{\mathrm{motion}}(X_t; X_{\mathrm{ref}}) \right)\rd t
\right].
\end{equation}
Crucially, these solutions apply to the general SOC formulation itself, independently of the specific choice of SOC solver.
In practice, we solve \cref{eq:method:final_obj} using Adjoint Matching (AM)~\citep{domingo2024am}, which is the efficient control learning algorithm.
The overall training procedure using AM loss is summarized in~\Cref{alg:am}.
\begin{figure}[t]
    \centering
    \includegraphics[width=\linewidth]
    {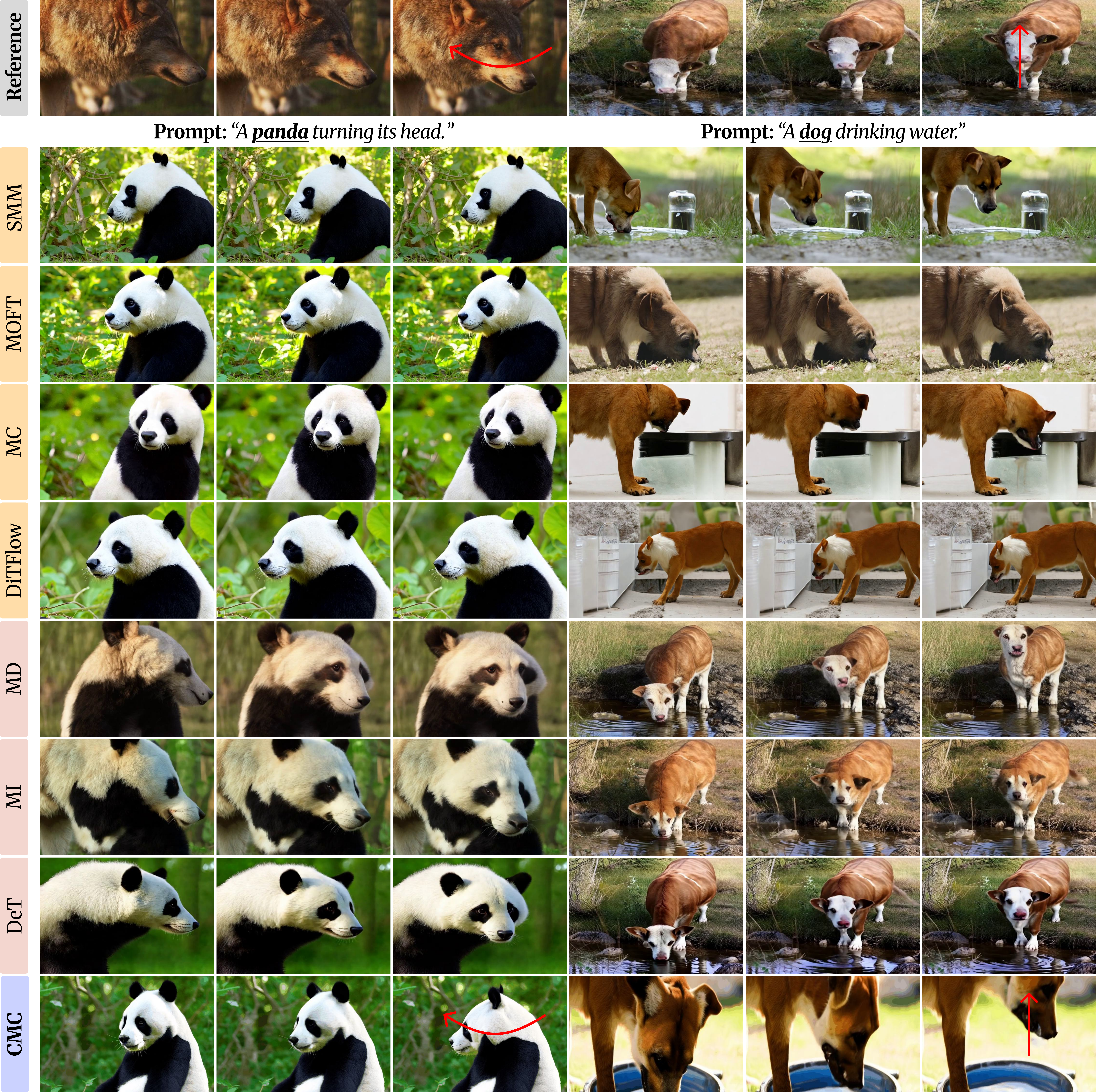}
    \vspace{-.4cm}
    \caption{\textbf{Qualitative comparison with baselines.} CMC achieves precise motion customization while significantly mitigating content leakage. The red arrow indicates the movement direction.}
    \vspace{-.4cm}
    \label{fig5:main_qual}
\end{figure}

\section{Experiments}
\label{sec:experiment}

\subsection{Experimental Setup}

\textbf{Evaluation Protocol.} We evaluate on reference videos sourced from DAVIS~\citep{perazzi2016} and prior work~\citep{pondaven2025video, ling2024motionclone, zhao2024motiondirector, maeffivmt}, paired with prompts of varying descriptive complexity.
Following the prior work, we report \emph{Motion Fidelity}~\citep{yatim2024}, \emph{Text Fidelity}, and \emph{Temporal Consistency}~\citep{huang2024vbench}. We additionally report \emph{Diversity} as an indirect measure of content leakage.
To evaluate diversity, we generate multiple trajectories per video–prompt pair, resulting in a total of 150 motion-customized videos.
We compare against seven baselines covering both training-free and tuning-based training schemes, and all baselines are adapted to CogVideoX-2B for a fair comparison.
More details on dataset, baselines, and metrics are in Appendix \S\ref{sec:appendix:exp_details}.

\textbf{Implementation Details.} We use CogVideoX-2B~\citep{yangcogvideox} as our base model with 28 denoising steps. The maximum timestep for the running cost ($\tau$ in continuous-time formulation) is set to 10, using features from the 15th transformer block, with $w(t)$ following a decaying schedule to avoid drastic change of dynamic. Full settings are in Appendix \S\ref{sec:appendix:imple_details}.

\definecolor{groupbg}{HTML}{F5F5F5}   % very very light gray for group header
\definecolor{trainigfreebg}{HTML}{FFEFD7}
\definecolor{tuningbasedbg}{HTML}{F6E9E6}
\definecolor{oursbg}{HTML}{C8CFFC}
\definecolor{ours}{HTML}{B3BCFF}

\begin{table}[t]
  \vspace{-0.1cm}
  \caption{Quantitative evaluation on CMC and baselines.}
  \label{tab:quantitative} 
  \vspace{-0.2cm}
  \footnotesize
  \setlength{\tabcolsep}{4pt}
  \renewcommand{\arraystretch}{0.95}
  \centering 
  \begin{tabular}{lccc!{\vrule width 0.8pt}c} 
  \toprule 
  \textbf{Method} & 
  \textbf{\makecell{Motion\\Fidelity $\downarrow$}}  &
  \textbf{\makecell{Text\\Fidelity $\uparrow$}} & 
  \textbf{\makecell{Temporal\\Consistency $\uparrow$}} & 
  \textbf{Diversity $\uparrow$} \\ 
  \midrule 
  {Backbone} & 0.2605 & 0.8520 & 0.9797 & 0.3139 \\
  \midrule 

  \rowcolor{trainigfreebg}\multicolumn{5}{l}{\textit{Training-free methods}} \\ 
  SMM~\citep{yatim2024}  & 0.1123 & 0.8229 & 0.9650 & 0.2457 \\ 
  MOFT~\citep{xiao2024video} & 0.1186 & 0.8272 & 0.9563 & 0.2292 \\ 
  MotionClone (MC)~\citep{ling2024motionclone} & 0.1303 & 0.7698 & 0.9253 & 0.2489 \\ 
  DiTFlow~\citep{pondaven2025video} & 0.1189 & 0.8304 & \textbf{0.9814} & 0.2402 \\ 

  \midrule 

  \rowcolor{tuningbasedbg}\multicolumn{5}{l}{\textit{Tuning-based methods}} \\ 
  MotionDirector (MD)~\citep{zhao2024motiondirector} & \textbf{0.0781} & \underline{0.8396} & 0.9566 & \underline{0.2623} \\ 
  MotionInversion (MI)~\citep{wang2025motion} & 0.0828 & 0.7991 & 0.9603 & 0.2206 \\ 
  DeT~\citep{shi2025decouple} & \underline{0.0821} & 0.8315 & 0.9667 & 0.2584 \\ 

  \rowcolor{oursbg} \textbf{CMC (Ours)} & 0.1093 & \textbf{0.8452} & \underline{0.9754} & \textbf{0.3243} \\ 
  \bottomrule 
  \end{tabular}
  \vspace{-12px}
\end{table}
\subsection{Main Results}

\textbf{Quantitative Comparison.}
\Cref{tab:quantitative} compares with the baselines, including the \emph{Backbone}, generated from the text prompt alone as a reference.
An ideal method would improve motion fidelity while matching the Backbone on the other metrics.
CMC achieves competitive motion fidelity while outperforming all baselines in both text fidelity and diversity, demonstrating that it generates diverse, motion-aligned samples that faithfully reflect the text prompt.
Although tuning-based baselines exhibit slightly higher motion fidelity, this stems from overfitting to the reference video, as detailed in~\cref{figS2} in Appendix \S\ref{sec:appendix:more_results}.
CMC surpasses all training-free methods in motion fidelity, striking a favorable trade-off between motion fidelity and diversity.

\begin{wrapfigure}{r}{0.5\textwidth}
    \centering
    \vspace{-15px}
    \includegraphics[width=0.5\textwidth]{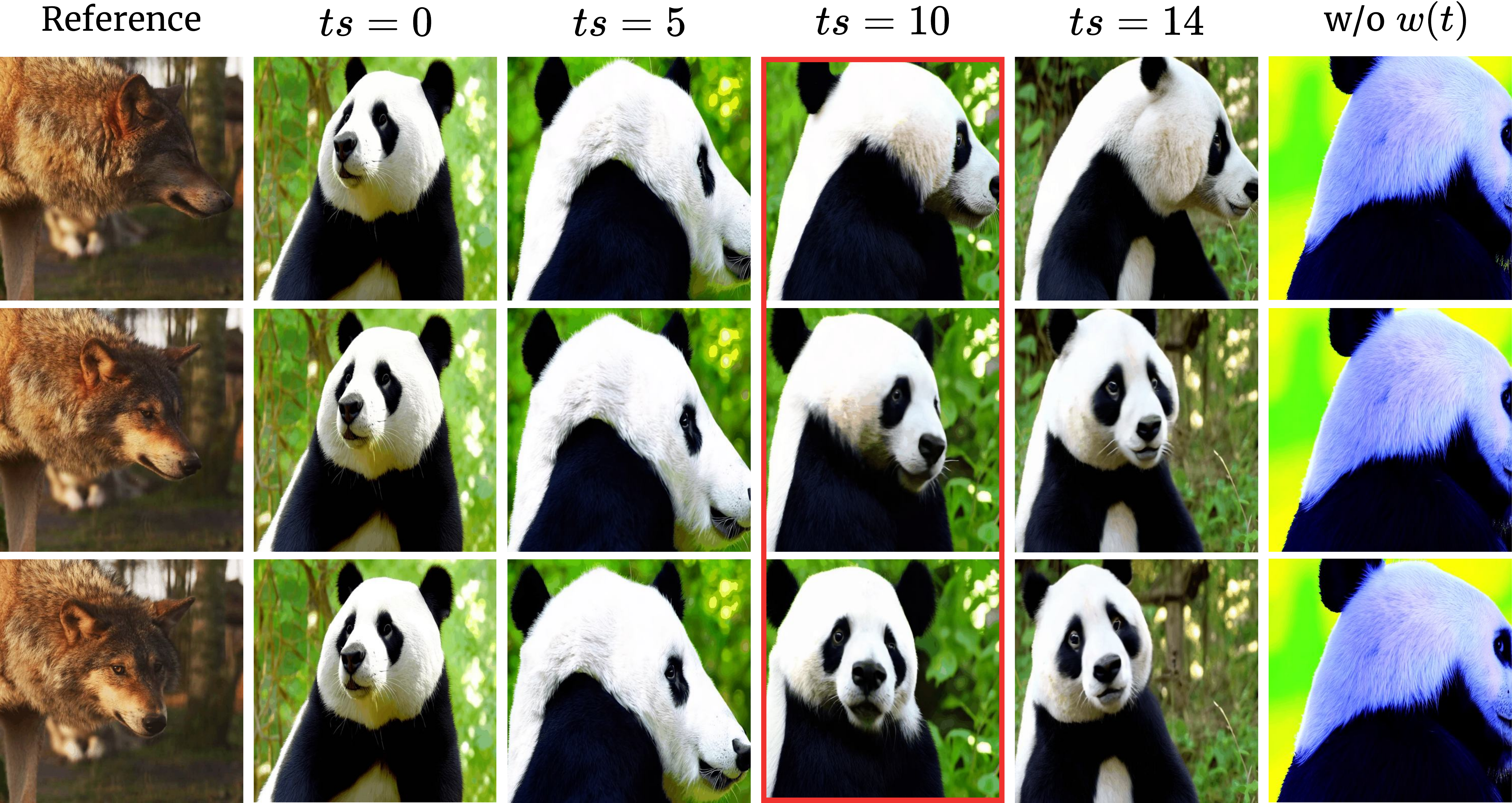}
    \vspace{-0.6cm}
    \caption{\textbf{Ablation study.} $ts$ denotes the motion injection timestep ($ts=0$ corresponds to the base T2V). Motion aligns with the reference from $ts=10$. Removing $w(t)$ leads to degenerated outputs.}
    \label{fig6:ablation}
    \vspace{-10px}
\end{wrapfigure}

\textbf{Qualitative Results.}
\cref{fig5:main_qual} compares CMC against baselines. Training-free methods produce plausible videos but fail to follow the reference motion. Tuning-based methods follow the motion well but suffer from content leakage, inheriting the layout, appearance, and color of the reference video. In contrast, CMC accurately transfers the reference motion while generating content faithful to the prompt, without collapsing toward the reference. Additional results are in Appendix \S\ref{sec:appendix:more_results}.

\textbf{Ablation Studies.} 
We investigate the impact of the\textbf{ maximum motion injection timestep $ts$} ($\tau$ in continuous-time formulation) to identify the optimal interval for motion guidance. As shown in~\Cref{fig6:ablation}, guidance restricted to very early timesteps ($ts=5$) is insufficient, while extending beyond $ts=10$ yields limited improvement but increases computational cost. We thus choose $ts=10$ as a balance between performance and efficiency. We further illustrate the effect of the \textbf{scaling factor} $w(t)$ in the running cost. Removing the schedule (constant weight) degrades sample quality, confirming that abrupt transitions between controlled and uncontrolled dynamics lead to distribution mismatch. The proposed decaying $w(t)$ mitigates this by ensuring smooth transitions. See Appendix \S\ref{sec:appendix:more_results} for more results and details.

\vspace{-0.1cm}
\section{Conclusion}
\label{sec:conclusion}

We present CMC, a principled framework for motion customization that significantly mitigates content leakage by preventing the generative process from collapsing toward the reference video. 
By formulating motion customization as a Stochastic Optimal Control (SOC) problem and tailoring it to the task, our approach enables precise, efficient and diverse motion-customized video generation.

\textbf{Limitations.} As our generation rely on strong expressivity of a pretrained model, producing outputs beyond its inherent capacity requires an additional reward model.
Also, while our experiments validate CMC against strong baselines, we leave generalization to arbitrary reference videos and prompts beyond the per-video optimization setting as future work, although we expect to transfer.% natural transferability.

\newpage
\section*{AI Use Statement}
We did not use generative AI tools for any tasks with required disclosure under
the ICLR 2027 policy. We used generative AI tools only to improve the readability
of the manuscript, such as polishing sentences and checking grammar. All
AI-assisted edits were reviewed by the authors, and we take full responsibility
for the content of this work.
\bibliography{iclr2027_conference}
\bibliographystyle{iclr2027_conference}

\newpage

\setcounter{page}{1}
\renewcommand{\thepage}{S\arabic{page}}
\setcounter{section}{0}
\setcounter{figure}{0}
\setcounter{table}{0}

\appendix
\clearpage
\hypersetup{
    colorlinks=true, % Turn on colored links
    linkcolor=black, % Set the link color to black
    filecolor=red, % Set the file link color to black
    urlcolor=blue, % Set the URL color to black
    pdftitle={Appendix}, % Optional: set PDF title
}
\setcounter{figure}{0}\renewcommand{\thefigure}{S\arabic{figure}}
\setcounter{table}{0}\renewcommand{\thetable}{S\arabic{table}}

\startcontents[appendix]
\section*{\textcolor{black}{Appendix}}
\printcontents[appendix]{}{1}{}\setcounter{tocdepth}{2}
\label{sec:appendix}

% ============================
\section{Stochastic Optimal Control}
\label{sec:appendix:soc}
We derive the induced terminal distribution in \cref{eq:method:reweighted_dist}.

\textbf{Setup.} Recall the base and controlled dynamics,
\begin{align}
\text{(base)}\quad
  & \mathrm{d}X_t = b(X_t,t)\,\mathrm{d}t + \sigma(t)\,\mathrm{d}W_t,
    && X_0\sim q_{\text{prior}}, \label{eq:appendix:base}\\
\text{(controlled)}\quad
  & \mathrm{d}X_t = \big[b(X_t,t) + \sigma(t)\,u(X_t,t)\big]\,\mathrm{d}t + \sigma(t)\,\mathrm{d}W_t,
    && X_0\sim q_{\text{prior}}, \label{eq:appendix:ctrl}
\end{align}
where $b(x,t) = f(x,t) + \sigma(t)^2 s^{\text{base}}(x,t)$. We denote by $p^{\text{base}}$ and $p^{u}$ the
path measures induced by \cref{eq:appendix:base} and \cref{eq:appendix:ctrl}. With $g\equiv 0$, the SOC
objective in \cref{eq:preliminary:soc_obj} becomes
\begin{equation}
J(u) = \mathbb{E}_{X\sim p^u}\!\left[\int_0^1\frac{1}{2} \|u(X_t,t)\|^2\,\mathrm{d}t + L(X)\right],
\qquad L(X) := \int_0^1 \ell(X_t,t)\,\mathrm{d}t.
\label{eq:appendix:obj}
\end{equation}

\textbf{Step 1: Control cost as KL divergence.} By the Girsanov theorem, the control cost is equivalent to the KL divergence between the controlled
process $p^u$ and the base process $p^{\text{base}}$, since both share the same initial distribution
$q_{\text{prior}}$:
$\mathrm{KL}(p^u\,\|\,p^{\text{base}}) = \mathbb{E}_{p^u}[\int_0^1\frac{1}{2}\|u(X_t,t)\|^2\,\mathrm{d}t]$.
Hence \cref{eq:appendix:obj} can be written purely in terms of path measures:
\begin{equation}
J(u) = \mathrm{KL}(p^u \,\|\, p^{\text{base}}) + \mathbb{E}_{X\sim p^u}\big[L(X)\big].
\label{eq:app_obj_kl}
\end{equation}

\textbf{Step 2: Optimal path measure.}
Define the tilted path measure
$\tilde{p}(X) := p^{\text{base}}(X)\exp\big(-L(X) + V(X_0,0)\big)$, where
$V(x,t) := -\log \mathbb{E}_{p^{\text{base}}}[\exp(-\int_t^1 \ell(X_s,s)\,\mathrm{d}s) \,|\, X_t = x]$
normalizes $\tilde{p}$ for each initial state $X_0$~\citep{domingo2024am}.
Then, for any path measure $p$ with initial marginal $q_{\text{prior}}$,
\begin{equation}
\mathrm{KL}(p \,\|\, p^{\text{base}}) + \mathbb{E}_{p}[L(X)]
  = \mathbb{E}_{p}\!\left[\log \frac{p(X)}{p^{\text{base}}(X)e^{-L(X)}}\right]
  = \mathrm{KL}(p \,\|\, \tilde{p}) + \mathbb{E}_{q_{\text{prior}}}\big[V(X_0,0)\big],
\end{equation}
which is minimized if and only if $p = \tilde{p}$, since the last term does not depend on $p$.
Therefore, the optimal path measure is the base path measure tilted by the accumulated running cost:
\begin{equation}
p^\star(X) = p^{\text{base}}(X)\,\exp\!\big(-L(X) + V(X_0,0)\big).
\label{eq:appendix:opt_path}
\end{equation}

% trajectory 분포를 우리가 실제로 관심 있는 X_1 = x 분포로 줄이기
\textbf{Step 3: Marginalizing to the terminal state.}
Factorizing $p^{\text{base}}(X) = p^{\text{base}}_1(X_1)\,p^{\text{base}}(X\,|\,X_1)$ in
\cref{eq:appendix:opt_path} and marginalizing over all trajectories ending at $X_1 = x$, we obtain
\begin{equation}
p^\star_1(x)
  = p^{\text{base}}_1(x)\;\mathbb{E}_{p^{\text{base}}}\!\left[\exp\!\Big(-\int_0^1 \ell(X_t,t)\,\mathrm{d}t + V(X_0,0)\Big) \,\Big|\, X_1 = x\right]
  = p^{\text{base}}_1(x)\,\phi(x),
\end{equation}
which is \cref{eq:method:reweighted_dist}. Consistently,
$\int p^{\text{base}}_1(x)\phi(x)\,\mathrm{d}x
 = \mathbb{E}_{q_{\text{prior}}}\big[e^{V(X_0,0)}\,\mathbb{E}_{p^{\text{base}}}[e^{-L(X)}\,|\,X_0]\big] = 1$,
so $p^\star_1$ is properly normalized.

{\setlength{\textfloatsep}{1pt}
 \setlength{\intextsep}{1pt}
\begin{algorithm}[t]
\caption{Control-based Motion Customization (CMC)}
\label{alg:am}
\begin{algorithmic}[1]
\Require Pre-trained score model $s^{\text{base}}$, step size $h$, fine-tuning iterations $N$
\State Initialize fine-tuned score model $s^\text{ft}$ from $s^\text{base}$ with parameters $\theta$
\For{$n \in \{0,\dots,N-1\}$}
    \State Sample $m$ trajectories $X = \{X_t\}_{t \in \{0,\dots,\tau\}}$ via controlled reverse SDE:
    \begin{equation}
    X_{t+h}
    =
    X_t
    +
    h \Big(
    f(X_t,t)
    +
    \sigma(t)^2 s^\text{ft}(X_t, t)
    \Big)
    +
    \sqrt{h}\,\sigma(t)\,\varepsilon_t,
    \;
    \varepsilon_t \sim \mathcal{N}(0,I),
    \;
    X_0 \sim \mathcal{N}(0,I)
    \label{eq:alg:reverse_sde}
    \end{equation}
    \State For each trajectory, solve the adjoint ODE backward from $t=\tau$ to $0$:
    \begin{equation}
    \begin{aligned}
    a_{t-h}
    &=
    a_t
    +
    h
    \Big(
    a_t^\top
    \nabla_{X_t}
    \big(
    f(X_t,t)
    +
    \sigma(t)^2s^{\text{base}}(X_t, t)
    \big)
    +
    \nabla_{X_t} \ell(X_t,t)
    \Big) \\
    a_\tau &= \nabla_{X_\tau} \ell(X_\tau,\tau)
    \end{aligned}
    \label{eq:alg:adjoint}
    \end{equation}
    \State For each trajectory, compute the Adjoint Matching objective on $[0,\tau]$:
    \begin{equation}
    \mathcal{L}_{\text{Trunc-AM}}(\theta)
    =
    \frac{1}{2}
    \sum_{t=0}^{\tau-h}
    \big\|
    \sigma(t)(s^\text{ft}-s^{\text{base}})
    +
    \sigma(t)^\top a_t
    \big\|^2
    \label{eq:alg:loss}
    \end{equation}
    \State Compute the gradient $\nabla_\theta \mathcal{L}(\theta)$ and update $\theta$
\EndFor
\State \Return Fine-tuned score model $s^{\text{ft}}$
\end{algorithmic}
\end{algorithm}
}
\section{Implementation Details}
\label{sec:appendix:imple_details}
\subsection{Motion Analysis Details}
For the analysis in \cref{fig4:early_quant}, we construct a controlled evaluation set using CogVideoX-5B as the base model. 
The dataset spans 10 subject categories (athlete, bear, cat, dog, human, kangaroo, monkey, panda, robot, and toddler) 
and 10 motion categories (backflipping, climbing, crawling, dancing, eating, punching, shaking hands, sitting down, spinning, and swimming). 
For each subject–motion pair, we generate 100 videos and extract intermediate noisy latents $X_t$ at each timestep, forming a latent dataset annotated with motion labels.
For every timestep, we train a separate 3-layer 3D CNN classifier to predict the motion category from the corresponding latent representation.
Evaluation is conducted on a held-out test set constructed with equal proportions from each subject–motion pair.

\subsection{Running Cost Design}
\label{sec:appendix:cost_design}
A straightforward approach to model motion is to use optical flow~\citep{teed2020raft}, which provides dense pixel-level motion vectors.
However, direct application of optical flow requires costly decoding process from latent to pixel space and is not available for noised latents. Therefore, we instead adopt a simple and effective instantiation that computes motion vectors directly in the feature space~\citep{geyer2023tokenflow, wang2024cove}.
Specifically, given a pair of frames $k,k' \in \{0, \dots, F-1\}$,
we define the feature extractor $\mathbf{F}:\mathbb{R}^n\times [0,1]\rightarrow\mathbb{R}^{H\times W\times C\times F}$ which extracts the feature map given latent $X_t \in \mathbb{R}^n$ and time $t$, and denote the $[i,j]$-coordinate patch in the $k$-th frame as  $\mathbf{F}(X_t,t)^{k}_{i,j}\in \mathbb{R}^C$.
Then, we define motion vector $\mathbf{v}:\mathbb{R}^n\times [0,1] \rightarrow \mathbb{R}^2$ of $\mathbf{F}(X_t,t)^{k}_{i,j}$ toward $k'(\neq k)$-th frame as
\begin{equation}
    \mathbf{v}(X_t,t)^{k,k'}_{i,j} = \underset{[i',j']}{\text{argmax}}\,d(\mathbf{F}(X_t,t)^k_{i,j},\mathbf{F}(X_t,t)^{k'}_{i',j'}) -[i,j], \quad [i',j']\in \Omega_{i,j},
\end{equation}
where $\Omega_{i,j}$ is a set of coordinates in local spatial window for $[i,j]$, and $d(\cdot, \cdot)$ is a similarity metric. $\mathbf{v}(X_t,t)^{k,k'}_{i,j}$ can be interpreted as optical flow of $[i,j]$ patch in $k$th frame heading to $k'$th frame in latent space. We define $\ell(x,t)$ as
\begin{equation}
    \ell(X_t,t) = w(t)\mathbb{E}_{i,j,k,k'}\|\mathbf{v}(X_t,t)_{i,j}^{k,k'}-\mathbf{v}(X_{\text{ref}},1)_{i,j}^{k,k'}\|^2_2,
\label{eq:our_running_cost}
\end{equation}
where $X_{\text{ref}}$ is an encoded latent of clean reference video and $w(t)$ is a scaling hyperparameter designed to mitigate the discontinuity caused by truncating control at $\tau$.

\begin{table}[t]
\centering
\caption{Training details of CMC.}
\label{tab:implementation}
\vspace{5pt}
\small
\renewcommand{\arraystretch}{0.9}
\setlength{\tabcolsep}{6pt}
\begin{tabular}{c|lc}
\toprule
\textbf{Category} & \textbf{Parameter} & \textbf{Value} \\
\midrule
\multirow{3}{*}{Model}
& Base model & CogVideoX-2B \\
& Resolution & $480 \times 720$ \\
& Denoising steps & 28 \\
\midrule
\multirow{4}{*}{LoRA}
& Rank & 32 \\
& Alpha & 32 \\
& Dropout & 0.1 \\
& Target modules & \texttt{to\_q},\hspace{2pt} \texttt{to\_k},\hspace{2pt} \texttt{to\_v},\hspace{2pt} \texttt{to\_out} \\
\midrule
\multirow{2}{*}{Training}
& Optimizer & AdamW \\
& Learning rate & $1 \times 10^{-5}$ \\
\midrule
\multirow{3}{*}{Control}
& Motion injection timestep $ts$  & 10 \\
& Cost weight $w(t)$
   & \cref{eq:appendix:cost_weight} \\
& Guidance scale & 6.0 \\
\midrule
\multirow{2}{*}{Cost}
& DiT block & 15 \\
& Local window & $9 \times 9$ \\
\bottomrule
\end{tabular}
\end{table}
\subsection{Training Algorithm}
Among the existing approaches to SOC, we adopt Adjoint Matching (AM)~\citep{domingo2024am}. AM offers an efficient and theoretically grounded objective by regressing the current control $u$ onto the lean adjoint state. Nevertheless, standard AM retains a key computational bottleneck of conventional SOC: evaluating its objective requires a full forward simulation from $t=0$ to $t=1$ for the terminal cost, followed by a backward adjoint simulation from $t=1$ to $t=0$. Our CMC formulation removes this bottleneck and yields a substantially more efficient AM variant.

Specifically, (i) setting $g(x) = 0$ eliminates the explicit terminal reward, which sets the terminal boundary condition of the adjoint state to zero: $a(1;X) = \nabla_x g(X_1) = 0$. Next, (ii) restricting control to the initial timesteps ensures that the running cost $\ell(x, t)$ vanishes for $t \in (\tau, 1]$. Together, these two conditions guarantee via the adjoint ODE that $a(t;X) = 0$ for all $t \in (\tau, 1]$.
Now the control objective simplifies to:
\begin{equation}
\mathcal{L}(u)
:=\mathbb{E}_{p^{\bar u}}\left [
\frac{1}{2}
\int_0^\tau
\| u(X_t,t) + \sigma(t)^\top a(t;X) \|^2 \rd t\
\right ], \quad X\sim p^{\bar u},
\label{eq/sec3.4/am_loss_ours}
\end{equation}
where $\bar u = \text{stopgrad}(u)$, and adjoint state is defined as
\begin{align}
\frac{\rd}{\rd t} a(t;X)
&=
- \Big( a(t;X)^\top \nabla_x b(X_t,t)
+ \nabla_x \ell(X_t,t)\Big), \quad t\in[0,\tau],\; X\sim p^{\bar u},
\label{eq/sec3.4/adjoint_ode_ours}\\
 a(\tau;X)
&=
\nabla_x \ell(X_{\tau},\tau), \qquad
 a(t;X)=0,\ \forall\, t\in(\tau,1].
\label{eq/sec3.4/adjoint_initial_ours}
\end{align}

This implies that both the forward trajectory and the backward adjoint simulation need only be computed over the truncated interval $t \in [0, \tau]$, eliminating unnecessary computation over $(\tau, 1]$. The overall training procedure is summarized in \Cref{alg:am}.

\subsection{Hyperparameter Details}
Detailed training configurations and hyperparameter settings used throughout the experiments are provided in \cref{tab:implementation}. We maintain the baseline model's default settings as specified, with the exception of the denoising process, which is set to 28 steps across all models for a fair comparison.

\section{Experiment Details}
\label{sec:appendix:exp_details}
\subsection{Dataset Description} 
We use 10 reference videos sourced from DAVIS~\citep{perazzi2016} and prior works~\citep{pondaven2025video, ling2024motionclone, zhao2024motiondirector, maeffivmt}, encompassing both camera and object motion. Each video is paired with 3 prompts of varying descriptive complexity. To evaluate diversity, we generate 5 trajectories per video–prompt pair, resulting in a total of 150 motion-customized videos, comparable in scale to prior works.

\subsection{Baseline Description}
\textbf{Training-free methods.} SMM~\citep{yatim2024} introduces a space-time feature loss based on spatially averaged diffusion features, and optimizes the latents during sampling to preserve the reference motion. MOFT~\citep{xiao2024video} extracts motion-aware features by filtering motion-relevant channels of diffusion features, and uses them to guide the generation. MotionClone~\citep{ling2024motionclone} utilizes sparse temporal attention weights as motion representations to guide the generation process. DiTFlow~\citep{pondaven2025video} extracts Attention Motion Flow (AMF) from cross-frame attention maps and performs training-free latent optimization to reproduce the reference motion in DiTs.

\textbf{Tuning-based methods.} MotionDirector~\citep{zhao2024motiondirector} employs a dual-path LoRA architecture to decouple appearance and motion learning. MotionInversion~\citep{wang2025motion} learns temporal and appearance embeddings injected into the attention layers to represent the reference motion. DeT~\citep{shi2025decouple} introduces a shared temporal kernel that smooths DiT features along the temporal axis, together with a dense point tracking loss, to decouple and learn motion in DiTs.

\textbf{Implementation.} For SMM and MOFT, we use the DiT implementations provided by \citet{pondaven2025video}, where DDIM inversion in SMM is replaced with key-value injection. For MotionClone, since CogVideoX has no dedicated temporal attention, we extract the cross-frame attention among tokens at the same spatial location from the 3D full attention, renormalized over frames, and apply its sparse top-1 loss within the same latent optimization pipeline. MotionInversion is adapted to our DiT backbone following \citet{shi2025decouple}, by injecting temporal embeddings into the query and key and spatial embeddings into the value of the 3D full attention. DiTFlow and DeT are natively built on CogVideoX, so we use its official implementation. For MotionDirector, since CogVideoX lacks separate spatial and temporal modules, we attach both the spatial and temporal LoRAs to the same attention and feed-forward layers of all DiT blocks, and distinguish them by their training data: the spatial LoRA is trained on single frames, while the temporal LoRA is trained on the full clip with the appearance-debiased loss, with the spatial LoRA kept frozen. At inference, only the temporal LoRA is applied, and DDIM inversion is replaced with key-value injection following \citet{pondaven2025video}.

\subsection{Metric Description}
\textbf{Text Fidelity} measures how accurately the generated video captures the \emph{non-motion content} specified in the text prompt. While standard metrics like CLIP score evaluate overall semantic correspondence, they can be sensitive to motion-related discrepancies that are inherent in motion customization tasks. Instead, we focus on object-level grounding to provide a more direct and interpretable measure of content fidelity, as our prompts primarily target specific objects.
To this end, we utilize the confidence score of Grounding DINO~\citep{liu2024groundingdino} as our metric. Specifically, we extract the target object noun from the prompt and calculate the average of the maximum confidence scores across all video frames. A higher score signifies that the target object is consistently and clearly present, indicating successful content reflection without interference from the reference video's appearance attributes.

\textbf{Motion Fidelity} measures how well the motion characteristics of the reference video are preserved in the generated video, following prior motion transfer evaluation protocols~\citep{yatim2024}. Specifically, point tracklets are extracted from both input and output videos using off-the-shelf point tracking method~\citep{cotracker3}, and motion fidelity is quantified by computing a Chamfer-style distance between the two sets of tracklets, where lower values indicate better motion preservation.

\textbf{Temporal Consistency} is evaluated using VBench~\citep{huang2024vbench}, a comprehensive video evaluation benchmark. It measures the overall consistency of generated videos by averaging the scores of subject consistency and background consistency.

\textbf{Diversity} serves as an indirect measure of content leakage; if a model overfits to the reference, it tends to produce nearly identical samples. Conversely, a robust model should retain the stochastic nature of the base text-to-video model, yielding visually distinct results for the same reference-prompt pair. To quantify this, we generate $N$ videos for a fixed input pair and represent each video as a flattened pixel-space vector $\mathbf{x}_i \in \mathbb{R}^d$. Diversity is then defined as the mean $\ell_2$ distance of these samples from their centroid $\bar{\mathbf{x}}$:
\begin{equation}
\text{Diversity} = \frac{1}{N}\sum_{i=1}^{N} |\mathbf{x}_i - \bar{\mathbf{x}}|^2, \quad \text{where} \quad \bar{\mathbf{x}} = \frac{1}{N}\sum_{i=1}^{N}\mathbf{x}_i.
\end{equation}
This parameter-free geometric measure provides an interpretable proxy for the model's output variance. A higher score indicates that the model is not merely replicating the reference video, but is instead generating novel scenes while capturing the underlying motion.

\section{More Results}
\label{sec:appendix:more_results}
\subsection{More Quantitative Results}

\textbf{Motion Analysis.} To examine whether early motion formation generalizes across model architectures and scales, we reproduce our motion-formation probing experiment (\Cref{fig4:early_quant}, based on CogVideoX-5B) on Wan2.2-5B and Wan2.2-14B~\citep{wan2025wan} using a smaller-scale setup. As shown in \Cref{tab:motion_formation_scale}, all models exhibit the same qualitative trend: probing accuracy rises sharply within the early denoising steps and saturates thereafter, confirming that motion structure consistently emerges early in the denoising process regardless of model scale.
\begin{table}[t]
\centering
\caption{Motion formation timestep across model scales.}
\label{tab:motion_formation_scale}
\begin{tabular}{lcccccc}
\toprule
\multirow{2}{*}{Timestep} & \multicolumn{2}{c}{CogVideoX-5B} & \multicolumn{2}{c}{Wan2.2-5B} & \multicolumn{2}{c}{Wan2.2-14B} \\
\cmidrule(lr){2-3} \cmidrule(lr){4-5} \cmidrule(lr){6-7}
 & Acc (\%) & $\Delta$ & Acc (\%) & $\Delta$ & Acc (\%) & $\Delta$ \\
\midrule
0  & 10.31 & -- & 12.00 & --     & 10.90 & --     \\
10 & 83.06 & +72.75 & 39.41 & +27.41 & 69.10 & +58.20 \\
20 & 93.16 & +10.10 & 47.06 & +7.65  & 91.00 & +21.90 \\
30 & 96.33 & +3.17 & 52.47 & +5.41  & 94.10 & +3.10  \\
40 & 96.53 & +0.20 & 55.18 & +2.71  & 96.30 & +2.20  \\
50 & 98.27 & +1.74 & 55.29 & +0.11  & 96.00 & --0.30 \\
\bottomrule
\end{tabular}
\end{table}

\textbf{Running Cost.} 
We argue that our framework's robustness to content leakage is structural rather than from any particular choice of running cost $\ell(x,t)$. This implies that our formulation remains inherently robust to content leakage even with a naive, appearance-entangled cost; unlike prior methods that rely heavily on carefully disentangling motion from appearance.

To verify this, we instantiate CMC with two alternative running costs:
\begin{enumerate}[leftmargin=*, itemsep=1pt, topsep=1pt]
    \item \emph{Simple feature-matching}, which directly matches
    the raw output feature of the 15th transformer block between the generated and reference videos without any explicit motion abstraction, evaluated both as a training-free sampling-time guidance and as a training-based fine-tuning objective.
    \item \emph{MotionClone-style cost}, which
    matches the 15th-block temporal attention using the sparse top-1 objective of MotionClone~\citep{ling2024motionclone}, adapted to our DiT backbone.
\end{enumerate}
Because raw features still entangle appearance information, both are deliberately appearance-mixed costs that would be expected to leak strongly if applied without our framework.

\begin{table}[H]
\centering
\caption{Effect of CMC on top of existing methods.}
\label{tab:cmc_on_existing}
\resizebox{\linewidth}{!}{%
\begin{tabular}{lcccc}
\toprule
\multirow{2}{*}{Running cost} & \multicolumn{2}{c}{Motion Fidelity $\downarrow$} & \multicolumn{2}{c}{Diversity $\uparrow$} \\
\cmidrule(lr){2-3} \cmidrule(lr){4-5}
 & w/o CMC & w/ CMC (Ours) & w/o CMC & w/ CMC (Ours) \\
\midrule
Simple feature-matching (training-free) & 0.145 & \textbf{0.140} & 0.197 & \textbf{0.206} \\
Simple feature-matching (training-based) & \textbf{0.134} & - & 0.171 & - \\
MotionClone-style cost & 0.188 & 0.212 & 0.254 & \textbf{0.282} \\
\bottomrule
\end{tabular}%
}
\end{table}
\Cref{tab:cmc_on_existing} reports motion fidelity and diversity for each cost, with and without CMC. Applying CMC on top of either existing cost consistently improves diversity (\textit{e.g.},
$0.197 \rightarrow 0.206$ for training-free feature-matching, and $0.254 \rightarrow 0.282$ for the
MotionClone-style cost), while motion fidelity remains comparable. In contrast, training-based feature-matching variant achieves the best motion alignment ($0.134$), but at the cost of the lowest diversity ($0.171$), which is a clear sign of content leakage. Fine-tuning directly on appearance-entangled features inevitably causes the model to overfit to the reference appearance. 
In summary, these results confirm that leakage prevention in CMC is structural rather than cost-dependent, ensuring robustness across diverse running cost designs.

\begin{figure}[t]
    \centering
    \includegraphics[width=\linewidth]
    {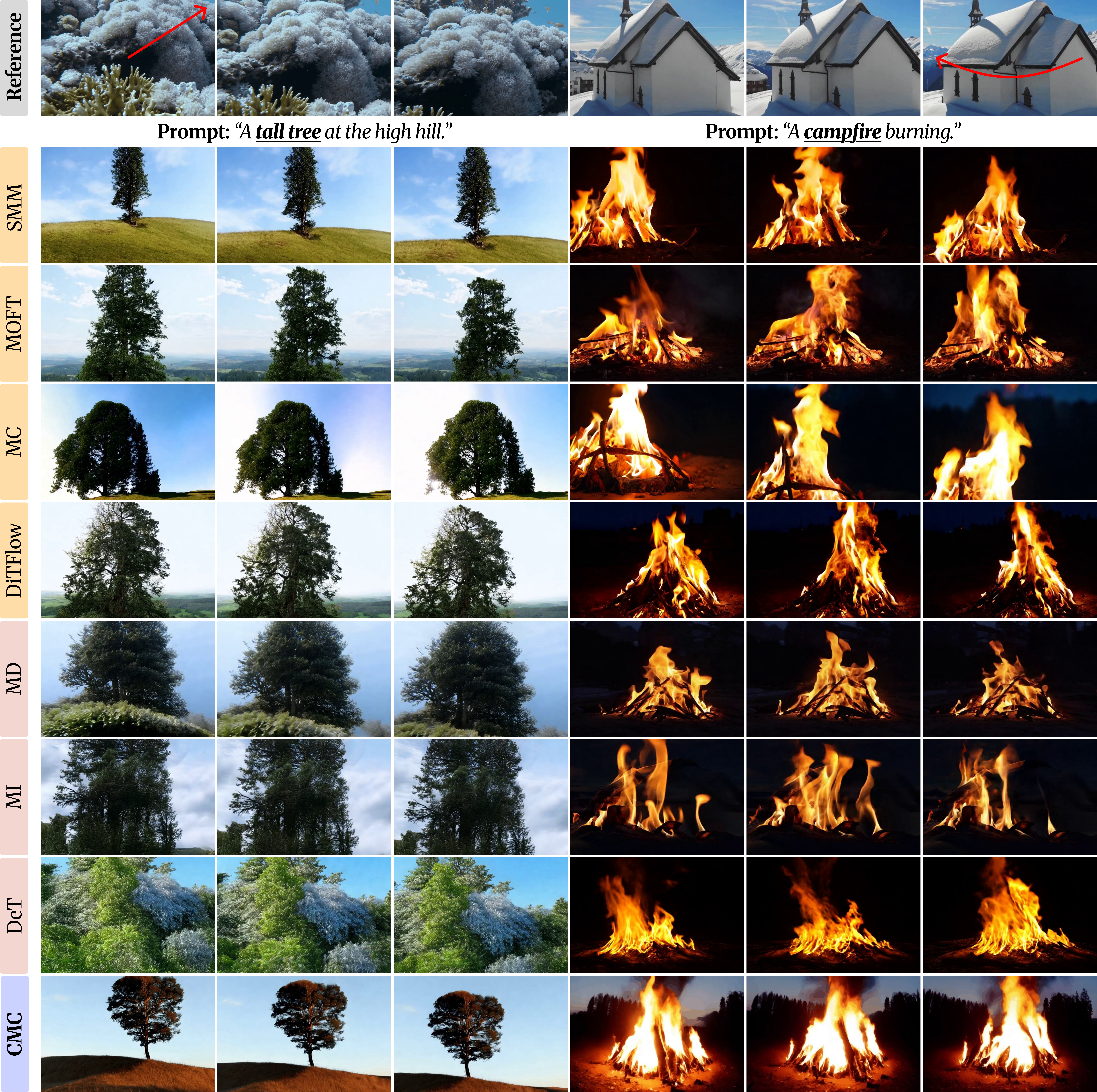}
    \vspace{-.4cm}
    \caption{\textbf{Additional qualitative comparison with baselines.} The red arrow indicates the camera movement direction.}
    \label{figS1}
    \vspace{-.4cm}
\end{figure}

\begin{figure}[t]
    \centering
    \includegraphics[width=\linewidth]
    {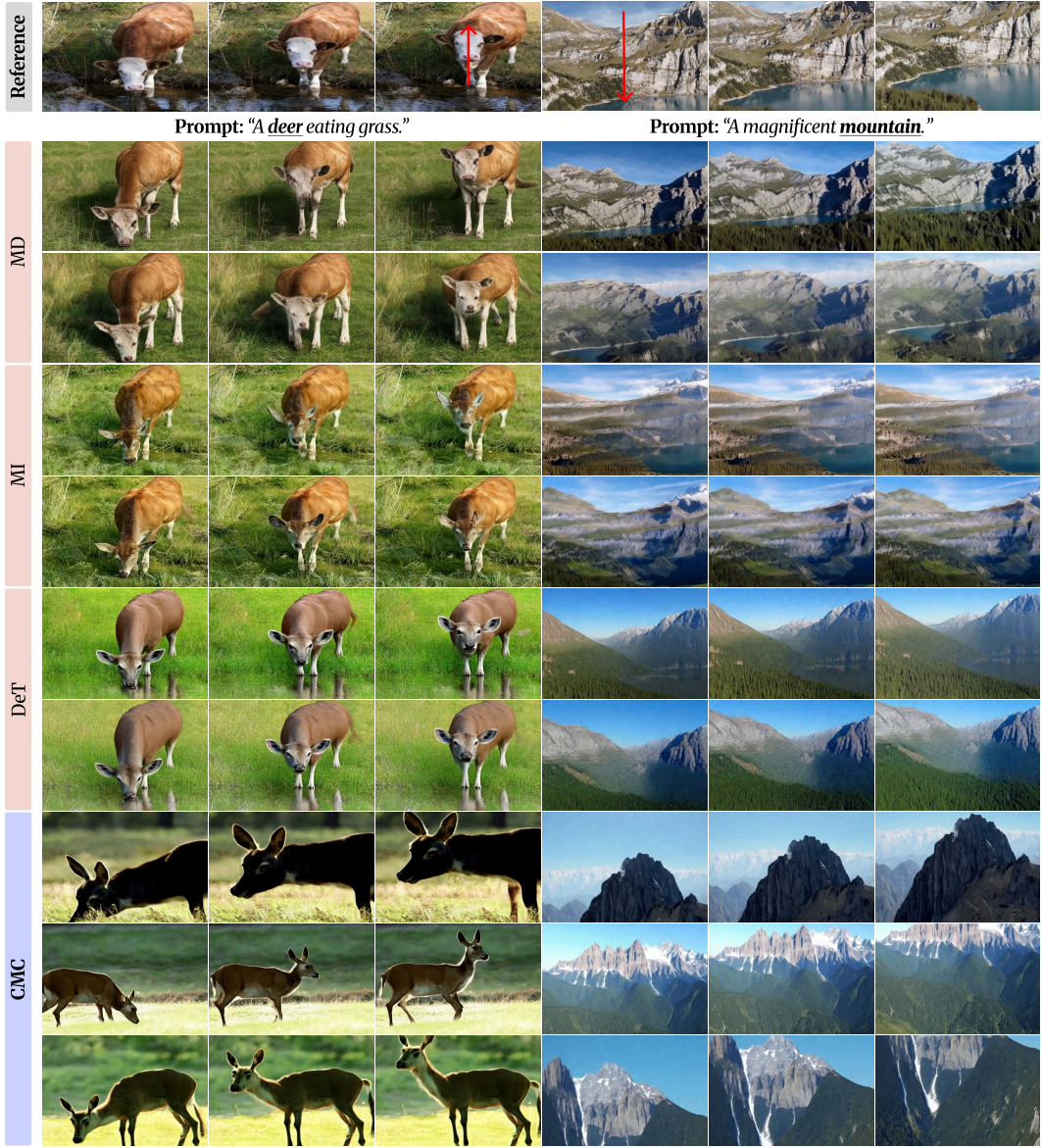}
    \vspace{-.4cm}
    \caption{\textbf{Comparison with tuning-based methods.} Tuning-based methods suffer from content leakage, producing near-identical samples that resemble the reference video, whereas CMC generates diverse samples. The red arrow indicates the movement direction.}
    \label{figS2}
    \vspace{-.4cm}
\end{figure}

\subsection{More Qualitative Results}
We provide additional qualitative results to further demonstrate our framework's capability.

\begin{figure}[t]
    \centering
    \includegraphics[width=\linewidth]
    {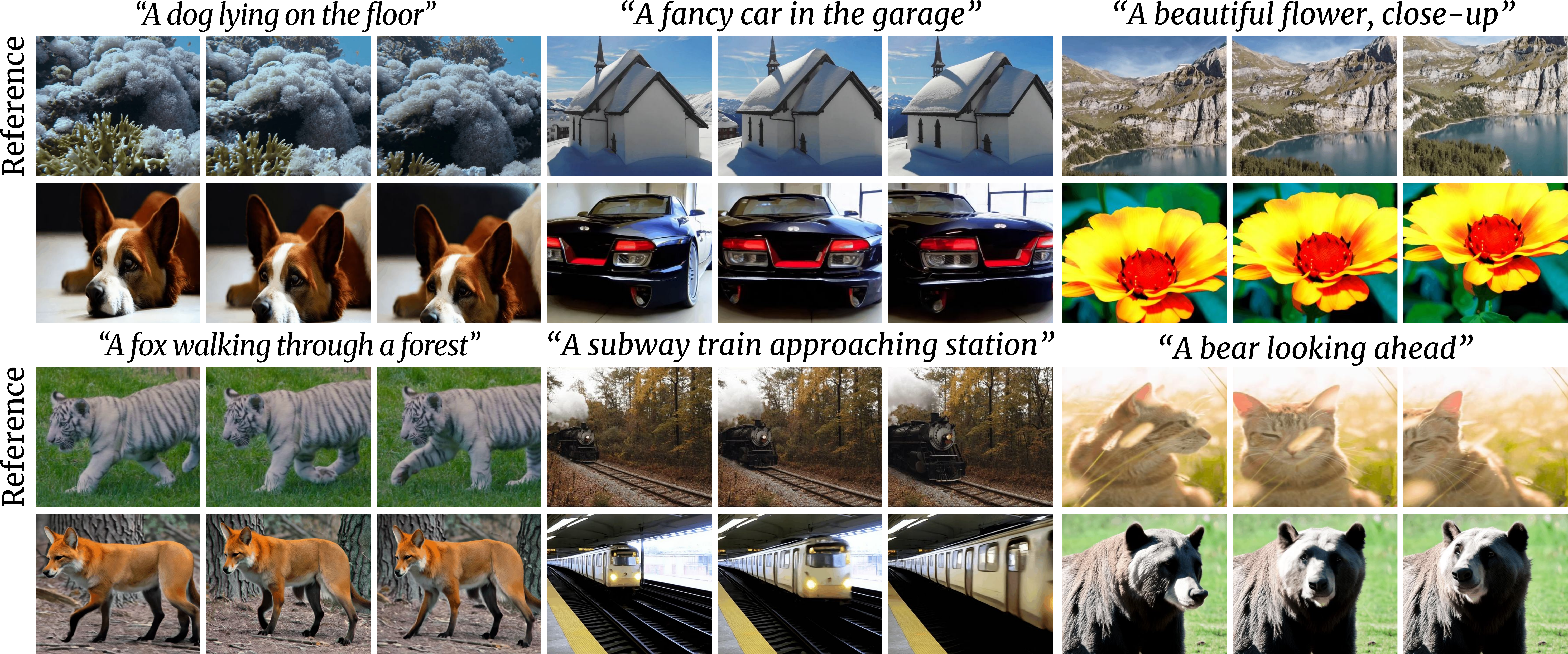}
    \vspace{-.4cm}
    \caption{More qualitative results of CMC.}
    \label{figS3}
    \vspace{-.4cm}
\end{figure}

\textbf{Visual Results.} \cref{figS1} presents additional comparisons, including camera motion. As in the main results, training-free methods fail to follow the reference motion and tuning-based methods suffer from content leakage, whereas CMC faithfully transfers the motion while reflecting the prompt. \cref{figS2} shows that tuning-based methods produce near-identical samples due to content leakage, while CMC produces diverse ones. \cref{figS3} presents a gallery of diverse CMC results.

\begin{figure}[t]
    \centering
    \includegraphics[width=\linewidth]
    {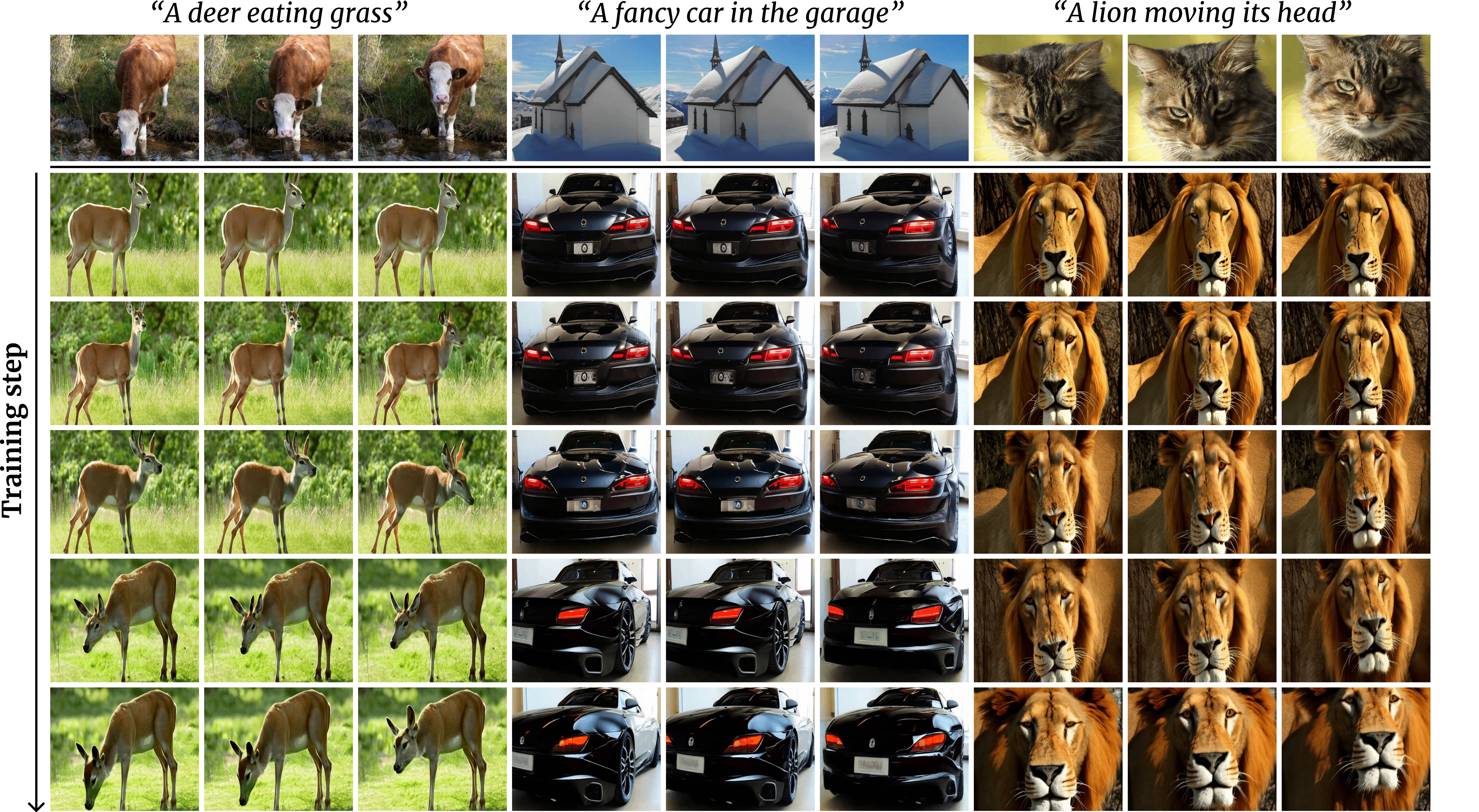}
    \vspace{-.4cm}
    \caption{Training behavior of CMC.}
    \label{figS4}
    \vspace{-.4cm}
\end{figure}

\textbf{Training Behavior.} \cref{figS4} visualizes the evolution of the base T2V sample throughout training. As training progresses, the generated sample gradually adapts to incorporate the reference motion, 
illustrating how the generative trajectory is progressively controlled toward the desired motion pattern. Importantly, this transition occurs in a stable manner, without abrupt structural changes, 
indicating a smooth adaptation of the generative process.

\subsection{More Ablation Results}

\textbf{Choice of \textit{ts}.}
In addition to the qualitative analysis in \cref{fig6:ablation}, we evaluate the impact of the maximum motion injection timestep $ts$ on computational efficiency and quantitative metrics.
As detailed in \cref{tab:further_ablations}(a), both computational overhead and memory requirements escalate rapidly as $ts$ increases. Notably, our primary setting of $ts = 10$ offers a significant efficiency advantage, operating approximately 2.5$\times$ faster than utilizing the full denoising sequence. At the same time, $ts = 10$ achieves the best motion fidelity with competitive diversity. This highlights that our approach not only preserves motion fidelity but also achieves a practical balance between performance and resource consumption. Experiments are conducted on a single a6000 gpu.

\textbf{Choice of $\boldsymbol{w(t)}$.}
To determine the optimal cost weight design, we evaluate four distinct schedules: (1) a constant weight ($w(t)=1.0$), (2) a linear decay, (3) a hybrid schedule (constant for the first half, then linear decay), and (4) our full setting, which adds an iteration-wise cost decay to the hybrid schedule.
To ensure a granular comparison, we fix a reference-prompt pair (\textit{e.g.}, ``wolf turning head'' $\to$ ``panda turning head'') and average the metrics over 100 samples.

As shown in \cref{tab:further_ablations}(b), our full setting achieves the best overall performance.
Empirically, we observe that while a sufficiently large weight is necessary to initiate motion transfer at early timesteps, maintaining a high cost throughout the entire training process can lead to optimization instability and fast drifts (\cref{fig6:ablation}).
By combining the hybrid temporal schedule with an iteration-wise decay, we achieve a robust balance between precise motion alignment and generative stability. Our final weight $w(t)$ is defined as:
\begin{equation}
  w(t) = \lambda_0 \cdot \tilde{w}(t) \cdot \gamma(n),
\label{eq:appendix:cost_weight}
\end{equation}
where we use $\lambda_0 = 400$.
The term $\tilde{w}(t)$ represents the hybrid temporal schedule, and $\gamma(n)$ denotes the iteration-wise decay:
\begin{equation}
\tilde{w}(t) = \begin{cases} 
1 & t \le 5 \\
\frac{10-t}{5} & 5 < t \le 10
\end{cases}, \quad
\gamma(n) = \begin{cases} 
50 & n \le 30 \\
30 & 30 < n \le 50 \\
10 & 50 < n \le 100 \\
5  & 100 < n \le 200 \\
1  & n > 200
\end{cases}
\end{equation}

% \newpage
\textbf{Choice of DiT Block.}
We examine which DiT block is most effective for extracting the features used in our running cost $\ell(X_t, t)$.
As shown in \cref{tab:further_ablations}(c), Block 15 achieves the best performance in both motion fidelity and diversity. This indicates that intermediate layers capture the most discriminative motion representations without over-constraining the generative process. This optimal balance confirms that intermediate-depth features effectively shape the trajectory while preserving the model's inherent generative richness. Additionally, the ablation on the local window size is provided in \cref{tab:further_ablations}(d).

\begin{table}[t]
\centering
\caption{Further ablation studies.}
\label{tab:further_ablations}

\begin{subtable}[b]{0.55\textwidth}
    \centering
    \setlength{\tabcolsep}{4pt}
    \begin{tabular}{c|cc|cc}
    \toprule
    $\boldsymbol{ts}$ & \textbf{Training} & \textbf{Mem} & \textbf{MF $\downarrow$} & \textbf{Div $\uparrow$} \\ \midrule
    5  & 195.6min & 27.5GB & \underline{0.0343} & \textbf{0.2453} \\
    \textbf{10} & 348.7min & 30.9GB & \textbf{0.0335} & \underline{0.2331} \\
    20  & 682.0min & 37.4GB & 0.0392 & 0.2198 \\
    28 & 944.6min & 42.4GB & 0.0421 & 0.2071 \\
    \bottomrule
    \end{tabular}
    \caption{Motion injection timestep}
\end{subtable}
\\[10pt]

\begin{subtable}[b]{0.30\textwidth}
    \centering
    \setlength{\tabcolsep}{4pt}
    \begin{tabular}{c|cc}
    \toprule
    $\boldsymbol{w(t)}$ & \textbf{MF $\downarrow$} & \textbf{Div$\uparrow$} \\
    \midrule
    Constant & 0.033 & \textbf{0.3162} \\
    Linear & 0.061 & 0.2446 \\
    Hybrid & 0.090 & 0.2389 \\
    \textbf{Ours} & \textbf{0.019} & \underline{0.2515} \\
    \bottomrule
    \end{tabular}
    \caption{Cost weight schedule}
\end{subtable}
\hspace{10pt}
\begin{subtable}[b]{0.30\textwidth}
    \centering
    \setlength{\tabcolsep}{4pt}
    \begin{tabular}{c|cc}
    \toprule
    \textbf{Block} & \textbf{MF $\downarrow$} & \textbf{Div$\uparrow$} \\
    \midrule
    10  & 0.0343 & 0.2180 \\
    \textbf{15} & \textbf{0.019} & \textbf{0.2515} \\
    20  & 0.0878 & 0.2210 \\
    30  & 0.0323 & 0.2199 \\
    \bottomrule
    \end{tabular}
    \caption{DiT Block}
\end{subtable}
\hspace{10pt}
\begin{subtable}[b]{0.30\textwidth}
    \centering
    \setlength{\tabcolsep}{4pt}
    \begin{tabular}{c|cc}
    \toprule
    \textbf{Window} & \textbf{MF $\downarrow$} & \textbf{Div$\uparrow$} \\
    \midrule
    None   & 0.1367 & 0.2501 \\
    (3, 3) & 0.1319 & 0.2468 \\
    (7, 7) & 0.0685 & 0.2505 \\
    \textbf{(9, 9)} & \textbf{0.019} & \textbf{0.2515} \\
    \bottomrule
    \end{tabular}
    \caption{Window size}
\end{subtable}
\end{table}

\end{document}